\documentclass[lettersize,journal]{IEEEtran}
\usepackage{amsmath,amsfonts}
\usepackage{algorithmic}
\usepackage{algorithm}
\usepackage{array}
\usepackage[caption=false,font=normalsize,labelfont=sf,textfont=sf]{subfig}
\usepackage{textcomp}
\usepackage{stfloats}
\usepackage{url}
\usepackage{verbatim}
\usepackage{graphicx}
\usepackage{cite}
\usepackage{url}
\usepackage{epstopdf}
\usepackage{booktabs}
\usepackage{placeins}
\usepackage{tikz}
\usepackage{subcaption}
\usepackage{amssymb, bm}
\begin{document}

\title{SolarCrossFormer: Improving day-ahead Solar Irradiance Forecasting by Integrating Satellite Imagery and Ground Sensors}

\author{Baptiste~Schubnel, 
        Jelena~Simeunović, 
        Corentin~Tissier, 
        Pierre-Jean~Alet 
        and~Rafael~E.~Carrillo
\thanks{The authors are with CSEM, Neuch\^{a}tel, Switzerland (e-mails: baptiste.schubnel@csem.ch, jelena.simeunovic@csem.ch, corentin.tissier@csem.ch, pierre-jean.alet@csem.ch, rafael.carrillo@csem.ch).}
}%

\markboth{Accepted for publication in IEEE Transactions on Sustainable Energy}%
{Schubnel \MakeLowercase{\textit{et al.}}: SolarCrossFormer}


\maketitle

\begin{abstract}
Accurate day-ahead forecasts of solar irradiance are required for the large-scale integration of solar photovoltaic (PV) systems into the power grid. However,  current forecasting solutions lack the temporal and spatial resolution required by system operators. In this paper, we introduce SolarCrossFormer, a novel deep learning model for day-ahead irradiance forecasting, that combines satellite images and time series from a ground-based network of meteorological stations. SolarCrossFormer uses novel graph neural networks to exploit the inter- and intra-modal correlations of the input data and improve the accuracy and resolution of the forecasts. It generates probabilistic forecasts for any location in Switzerland with a 15-minute resolution for horizons up to 24 hours ahead. One of the key advantages of SolarCrossFormer its robustness in real life operations. It can incorporate new time-series data without retraining the model and, additionally, it can produce forecasts for locations without input data by using only their coordinates. Experimental results over a dataset of one year and 127 locations across Switzerland show that SolarCrossFormer yield a normalized mean absolute error of 6.1 \% over the forecasting horizon. The results are competitive with those achieved by a commercial numerical weather prediction service.
\end{abstract}

\begin{IEEEkeywords}
Solar energy, solar radiation, forecasting, graph neural networks.
\end{IEEEkeywords}

\section{Introduction}
\IEEEPARstart{T}{he} growing capacity of solar power sources poses a challenge for distribution system operators, balance group managers and traders due to the inherent variability of solar power. Therefore, accurate short to medium-term forecasting of local solar production is essential \cite{Yang2022}. However, existing solutions often lack in spatial and temporal resolution at the forecasting horizon required by system operators.

Since global horizontal irrandiance (GHI) is the main factor influencing the power generation of photovoltaic (PV) plants, a large portion of existing solar forecasting works are dedicated to GHI forecasting, and the GHI forecasts are subsequently converted to PV power forecasts \cite{Yang2022}. Classical approaches for solar forecasting combine numerical weather predictions (NWP), satellite images and ground measurements with physical models \cite{Antonanzas2016}. These methods come with high computational and storage demands, thus they are often implemented with low temporal and spatial resolution. In contrast, data-driven solutions that rely solely on data from a network of ground-based sensors have shown state-of-the-art results for intra-day irradiance forecasting while requiring lower computational resources \cite{Khodayar2020,Simeunovic2022a,Simeunovic2022b,Carrillo2023, GHIMIRE2019113541}. Nevertheless, extending these solutions to longer horizons, e.g., day-ahead forecasts, entails providing additional information on cloud dynamics and a broader spatial context. For example, the authors of \cite{PERERA2024122971} used a hierarchical approach to fuse data from a network of on-site weather measurements and PV power production over a region to compute day-ahead forecasts of the regional PV power production. While effective, their approach was constrained to 18-hour horizons and aggregated outputs due to limited spatial coverage. This highlights a critical need for integrating satellite imagery, which offers wide-area observational data and cloud motion tracking, to complement ground-based measurements and enable more accurate, site-specific day-ahead solar forecasts. 

To addressed these challenges, recent advancements in solar forecasting have leveraged computer vision and deep learning techniques to enhance prediction accuracy \cite{Paletta2023}. By integrating multisensor earth observations, including data from sky cameras, satellites, and weather stations, researchers have improved real-time cloud cover analysis, a key factor in solar irradiance forecasting. Deep learning models capable of extracting relevant features from these diverse data sources have shown promising results for robust forecasting at single sites \cite{9324830, BRESTER2023266}. Methods for multi-modal data fusion can be divided by forecasting horizon: very-short-term horizons, that mainly use all sky images and ground data, short-term horizons, that fuses data from multiple sources, and day-ahead horizons, that mainly use data from satellite imagery and ground-based measurements.

Several studies have explored the fusion of all-sky images and ground-based measurements for very-short-term solar forecasting (minutes ahead). Ajith \textit{et al.} \cite{Ajith2021} combined infrared images with past irradiance data for local predictions, while Sarkis \textit{et al.} \cite{niu2025solar} used a lightweight transformer model integrating public camera images and historical GHI time-series. Paletta \textit{et al.} \cite{PALETTA2024118398} applied physics-informed transfer learning across locations, though their method relies heavily on physical models and diverse training data. Comparative analyses \cite{PALETTA2021855} show that incorporating spatio-temporal features from sky image sequences improves short-term accuracy, yet models still struggle with sudden weather changes \cite{9711179}. To address this, recent work has introduced vision transformers and attention mechanisms that better capture global cloud motion and fuse multimodal inputs, achieving more accurate forecasts up to one hour ahead \cite{LIU2023121160}.

Spatio-temporal fusion using satellite imagery and ground-based measurements has also been explored for short-term horizons (up to 4 hours). Paletta \textit{et al.} \cite{paletta2023omnivision} combined satellite and sky images using convolutional and recurrent networks as spatial and temporal encoders to improve forecasts up to 60 minutes ahead. Buzzi \textit{et al.} fused satellite images with local meteorological data to predict GHI for horizons up to 60 minutes ahead \cite{buzzi}. Models like IrradianceNet \cite{NIELSEN2021} and those by Carpentieri \textit{et al.} \cite{Carpentieri2023} use satellite images, optical flow and scale-dependent approaches to infer cloud dynamics and uncertainty for intra-day horizons.  Jing \textit{et al.} \cite{Jing2024} further extended this by fusing multichannel satellite and meteorological ground data using ConvLSTM and attention modules for regional forecasts. The authors of \cite{9176798} follow a similar approach using CNNs to extract cloud factors from satellite images, before fusion with optical flow and meteorological data. However, these models remain limited to intra-day horizons and often focus on single-site predictions.
 
In the context of PV power forecasting, similar trends are observed. Deep learning models combining satellite imagery and ground measurements have achieved promising results for short-term horizons (minutes to a few hours ahead) \cite{info14110617, en18071796, wang2024}, though most efforts mainly focus on forecasts for a single site. The works of Qin \textit{et al.} \cite{qin} and Attya \textit{et al.} \cite{attya} explored multi-site predictions using satellite-derived cloud motion and distributed sensors. Despite these advances, day-ahead forecasting across multiple sites using multimodal data remains largely unexplored. 

The potential of deep learning methods for day-ahead solar forecasting using satellite imaging data has been demonstrated in \cite{SEBASTIANELLI2024114431}. Recently, Boussif \textit{et al.} introduced the CrossVivit model \cite{Boussif2023}, which fuses satellite imaging data and time-series from the desired location to forecast irradiance. The model data leverages spatio-temporal context from satellite images to forecast irradiance at arbitrary locations. This approach was enhanced in SolarCube \cite{li2024solarcube}, which incorporated a more diverse dataset covering a broader range of weather conditions and offering higher temporal and spatial resolution. In addition to satellite and ground-based data, SolarCube also integrates physics-derived solar features to improve forecasting accuracy. Wang \textit{et al.} \cite{WANG2025119218} further adapted the CrossVivit architecture by introducing an improved satellite image encoder for day-ahead PV power forecasting.

Despite these advancements, existing models typically rely on data from a single location during inference, limiting their ability to capture broader spatio-temporal relationships between satellite imagery and distributed ground-based measurements. This presents a key opportunity: by fusing data from a dense network of ground sensors with satellite-based features, day-ahead solar forecasting can be enhanced, especially for multi-site operations.

To fill this gap, in this paper, we present SolarCrossFormer, a deep learning architecture for day-ahead irradiance forecasting (24 hours ahead horizon). SolarCrossformer extends previous works by some of the authors \cite{Simeunovic2022a,Carrillo2023} to day-ahead forecasting by including satellite imagery as an additional input and using novel graph neural networks (GNN) models to exploit the inter and intra-modal correlations of the two sensing modalities. Satellite images provide the wider spatial context of the cloud dynamics, while the ground measurements provide information on the local variations. The proposed model uses data of the past 24 hours from the two sensing modalities, without requiring numerical weather predictions as inputs, to generate probabilistic forecasts of irradiance. The main contributions of this paper are:
\begin{itemize}
\item Novel deep learning architecture: SolarCrossFormer fuses information from various sensing modalities at different spatial scales. By processing satellite images in a multiscale fashion and finding cross-relations between data from each sensing station and image patch, the model learns spatial and temporal features for forecasting across different horizons. This approach achieves the accuracy of intra-day methods for short-term forecasts and day-ahead methods for longer forecasts. 
\item Robustness and flexibility: The solution is independent of the number of input nodes in the sensing network, making it robust for real-life operations. Using a dynamic masking approach during training, SolarCrossFormer can incorporate time-series data from unseen locations without retraining the entire model. Additionally, it can generate forecasts for locations without input data by using their coordinates and leveraging data from existing sensor networks and satellite images.
\item Extensive evaluation: SolarCrossFormer was benchmarked against state-of-the-art approaches for day-ahead forecasting over a dataset of one full year and 127 locations distributed over Switzerland. The proposed approach was also compared against a commercial NWP solution for three months at Neuchâtel, Switzerland.
\end{itemize}

The organization of the paper is as follows. Section \ref{section_2} describes the problem formulation and the architecture of SolarCrossFormer model. In section \ref{section_3} we describe the experimental setup used for the evaluating the performance of the proposed model. We present the results of the evaluation of the proposed model in section \ref{section_4} and conclude in section \ref{section_5}.

\section{Methodology}
\label{section_2}
\subsection{Problem Formulation}
The task we address is to produce probabilistic forecasts of GHI for the next 24 hours on a set of $N_d$ desired locations given a sequence of $P$ past measurements from a network of weather sensors and past satellite images. We model the uncertainty of the predictions by computing $Q$ quantiles of the predictive distribution. 

Let $\{ \bm{x}_{ts}(t)\}_{t=t_0-T}^{t_0-1}$ and $\{ \bm{x}_{sat}(t)\}_{t=t_0-T}^{t_0-1}$ denote the sequence of $T$ past weather measurements and satellite images, respectively, where $\bm{x}_{ts}(t) \in \mathbb{R}^{N_t \times f}$ and $\bm{x}_{sat}(t) \in \mathbb{R}^{h \times w \times c}$. $N_t$ denotes the number of sensors (nodes) at time $t$ (possibly changing over time), $f$ the number of measured weather variables, $h$ and $w$ the high and width of the images in pixels, $c$ the number of spectral channels in the satellite data and $t_0$ defines the starting point of the forecast. The forecasting problem can be formulated as:
\begin{equation}
\label{forecast_problem}
\{ \hat{\mathbf{y}}(t)\}_{t=t_0}^{t_0+H-1} = f_{\mathbf{\theta}} \left( \{ \bm{x}_{ts}(t)\}_{t=t_0-T}^{t_0-1}, \{\bm{x}_{sat}(t)\}_{t=t_0-T}^{t_0-1} \right),
\end{equation}
where $\hat{\mathbf{y}}(t) \in \mathbb{R}^{N_d \times Q}$ denotes the array of $Q$ quantiles for the $N_d$ desired sites at lead time $t$, $H$ denotes the number of discrete time steps in the forecasting horizon and $f_{\mathbf{\theta}}(\cdot)$ is a parametric function with learnable parameters $\mathbf{\theta}$. In this work we focus on forecasting horizons of 24 hours with a resolution of 15 minutes, thus, $H=96$. We use a window of 24 hours for the past measurements which yields $T=96$.

\subsection{Architecture}
The neural architecture is built following ideas from the CrossFormer architecture \cite{zhang2023crossformer} and alternates temporal attention layers (self-attention for the time series) with pixels-nodes and nodes-nodes cross-attention layers to cross-correlate features between different sensor sources. It is directly inspired by \cite{Boussif2023} and extends information exchange via pixels-nodes and  nodes-nodes dot-product attention.  Unlike \cite{Boussif2023}, we did not use a vision transformer to extract features from the images, as it did not improve the model accuracy and led to computation overhead for high resolution images. 

\begin{center}
\begin{figure*}[th]
    \centering
    \includegraphics[scale=1.9, trim = 2 0 0 0, clip]{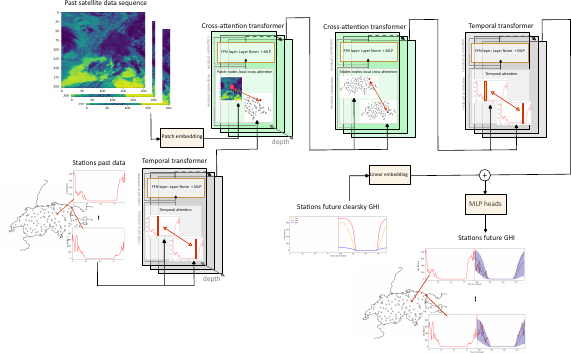} 
    \caption{SolarCrossFormer architecture with satellite data inputs. The encoder consists of a temporal transformer to encode the individual nodes' time series, a cross-attention transformer to correlate patch-node information and a second cross-attention transformer to correlate inter-node information. The decoder consists of a temporal transformer followed by a MLP.}
    \label{architecture}
\end{figure*}
\end{center}

The encoder-decoder architecture is depicted in Figure \ref{architecture}. The data representation flow for the encoder is the following. As in \cite{Boussif2023}, the embedded time series data from the weather data nodes first go through a temporal transformer  that computes  time correlation features for each node independently. A cross-attention transformer carries out the cross-correlation between the embedded pixels (patch embeddings) and the output of the temporal transformer. A second cross-attention transformer carries out the nodes-nodes correlation. Finally, the resulting representation is fed to the decoder that consists of a temporal transformer followed by a Multi Layer Perceptron (MLP) to map back to the prediction space feature dimension. The decoder also has as input the clear sky GHI data for the forecasted horizon. The clear sky GHI can be seen as a positional encoding that encodes seasonal and location information since it depends on the position of the sun with respect to earth at a particular time. We also implemented a version of SolarCrossFormer without satellite data, in which the first cross-attention layer is removed, and outputs from the time series transformer directly go to the node-nodes cross attention layer.

Each transformer layer uses the following ingredients: layer normalization \cite{ba2016layer}, scaled dot-product attention \cite{vaswani2017attention}, residual connection \cite{he2016deep} and feedforward networks. In the following, we give the detailed operations performed in the temporal attention and nodes-pixels cross-attention layers (Eqs. \eqref{linear} to \eqref{alph}). We  use Einstein summation convention for tensors (summation over repeated indices). We start defining the base operations for the attention layers. A  linear transformation $\mathcal{L}^{d \to \tilde{d}} : \mathbb{R}^{\dots \times d} \rightarrow  \mathbb{R}^{\dots \times \tilde{d} }$ is defined by
\begin{equation}
\label{linear}
(\mathcal{L}^{d \to \tilde{d}} \bm x) _{ \dots i }:= W_{ i j} \, \bm x_{ \dots j} + b_i,
\end{equation}
where $\bm{x} \in  \mathbb{R}^{\dots \times d}$ is the input tensor, with arbitrary first dimensions (represented with the $\dots$ symbol) up to the last layer of dimension $d$, $W_{ij}$ is a  weight matrix of dimension $\tilde{d} \times d$ and  $b_i$  is a bias vector of dimension $\tilde{d}$. A linear map with no bias is denoted here by $\mathcal{L}^{d \to \tilde{d}}_0$.
The  layer normalization map $\text{LayerNorm} : \mathbb{R}^{\dots \times d} \rightarrow  \mathbb{R}^{\dots \times  d  }$  is defined as 
\begin{equation}
\label{lnorm}
\text{LayerNorm}(\bm{x}): = \bm{\gamma} \odot \frac{\bm{x} - \mathbb{E}_d(\bm x)}{\sqrt{\mathbb{V}\text{ar}_d(\bm x) + \epsilon}} + \bm{\beta},
\end{equation}
where  $\mathbb{E}_d$ and  $\mathbb{V}\text{ar}_d$ are the mean and variance along the $d$ dimension, respectively,  $\bm{\beta}, \bm{\gamma}  \in \mathbb{R}^d$ are learnable parameters learnt during training, and  $\odot$ denotes the element-wise multiplication.  Finally, for three tensors $\bm{Q} \in \mathbb{R}^{\dots n \times d }, \bm{K} \in \mathbb{R}^{\dots m \times d }, \bm{V}  \in \mathbb{R}^{\dots m \times \tilde{d} } $, sharing the same dimensions over the $\dots$ indices, the scaled dot-product attention $\text{SA}:  \mathbb{R}^{\dots \times n \times  d} \times  \mathbb{R}^{\dots \times m \times  d} \times  \mathbb{R}^{\dots \times m \times  \tilde{d}} \rightarrow  \mathbb{R}^{\dots \times n \times \tilde{d}  } $ is defined by
\begin{equation}
\label{scaled_dot_prod}
\text{SA}(\bm{Q}, \bm{K}, \bm{V})_{\dots i h} =  \alpha_{ \dots ij}(\bm{Q}, \bm{K})  \bm{V}_{ \dots j h},
\end{equation}
where the attention weights $\alpha_{\dots ij}$ are computed as
\begin{equation}
\label{alph}
\alpha_{ \dots ij}(\bm{Q}, \bm{K}) := \frac{ \exp\left( \frac{\bm Q_{\dots i k} \bm  K_{\dots j k}}{\sqrt{d}} \right) }{ \sum_{j'} \exp\left( \frac{\bm Q_{\dots i k} \bm  K_{\dots j' k}}{\sqrt{d}} \right)  }.
\end{equation}
For transformers, a multi-head version with a distinct weight multiplication for each head  and each element $\bm{Q}, \bm{K}, \bm{V}$, followed by concatenation and projection is used in practice. We denote it by $\text{MSA}(\bm{Q}, \bm{K}, \bm{V})$.
\small
\begin{equation}
\begin{aligned}
\text{MSA}: (\bm{Q}, &\bm{K}, \bm{V}) \\
&\hspace{1.5em} \overset{\mathcal{L}_{0}^{d \to d_{\text{head}}}}{\underset{\times 3\, n_h}{\longmapsto}} 
\{(\bm{\tilde Q}^{(a)}, \bm{\tilde K}^{(a)}, \bm{\tilde V}^{(a)})\}_{a=1}^{n_h} \\
&\hspace{4.5em} \overset{\text{SA}}{\underset{\times n_h}{\longmapsto}} 
\{\bm{z}^{(a)}\}_{a=1}^{n_h} 
\overset{\text{Concat}}{\longmapsto} 
\bm{z} 
\overset{\mathcal{L}^{n_h d_{\text{head}} \to \tilde{d}}_0}{\longmapsto} 
\bm{y},
\end{aligned}
\label{MSA_eq}
\end{equation}
\normalsize
where $n_h$ denotes the number of heads. In the following we use these basic blocks to define the temporal transformer and cross-attention transformer layers used in the architecture.

\subsubsection{Temporal transformer ( Figure  \ref{fig:attn1}, right block)}
Before being fed to the temporal transformers, the ground stations time series data $\bm{x}_{ts} \equiv \{ \bm{x}_{ts}(t)\}_{t=t_0-P}^{t_0-1} \in \mathbb{R}^{N_t \times T \times  f}$  first undergo a sequence positional embedding;  see Figure \ref{architecture}. A cyclical encoding  ($\sin$ and $\cos$ functions applied to the minutes and hour coordinates of each measurement) is concatenated to the signal and then a linear embedding projects this concatenation to the embedding dimension $d$  from Table \ref{tab:parameters}. We denote the output of these two operations with the same notation $\bm{x}_{ts}  \in \mathbb{R}^{N_t \times T \times d }$. In the temporal transformer layer, self-attention is used  along the time axis (second to last index, see Eq. \ref{scaled_dot_prod}). The operations of the temporal transformer layer are:
\begin{align}
\bm{z}_0 & = \text{LayerNorm}(\bm{x}_{ts}),\nonumber \\
\bm{z}_1 &=   \bm{x}_{ts} +  \text{MSA}(\bm{z}_0,\bm{z}_0,\bm{z}_0), \nonumber\\
\bm{x}_{ts} &= \text{MLP}(\text{LayerNorm}(\bm{z}_1)) + \bm{z}_1, 
\end{align}
where $ \text{MLP}$ is a 2-layers feed forward network with Geglu activation function,  keeping the same dimension in output as in input, and wih hidden dimensions specified in Table \ref{tab:parameters}.
These operations are applied sequentially, repeated as many times as specified by the depth of the transformer architecture (Table \ref{tab:parameters}, Transformers depth).

\begin{figure}
\centering
\begin{minipage}{.25\textwidth}
  \centering
  \includegraphics[width=.65\linewidth]{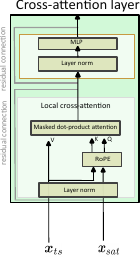}
\end{minipage}%
\begin{minipage}{.25\textwidth}
  \centering
  \includegraphics[width=.7\linewidth]{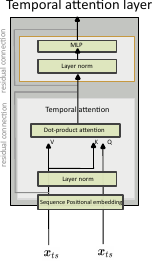}
\end{minipage}
\caption{Attention layers. Left: Cross-attention. Right: Temporal attention.}
\label{fig:attn1}
\end{figure}

\subsubsection{Transformer with cross-attention layer ( Figure  \ref{fig:attn1}, left block)}
The cross-attention layer is similar in spirit to the temporal attention layer, but involves cross-attention instead of self-attention, and requires transposing the indices to apply attention along the spatial indices.
We introduce in our work  a local dot-product masked attention to force the attention mechanism to focus on local information in 2D space. The sequence of satellite images $\bm{x}_{sat}  \in \mathbb{R}^{h \times w \times  T \times  c}$  first undergo a patch embedding, defined as:
\begin{equation*}
\underset{   \in \mathbb{R}^{h \times w \times  T \times  c}}{\bm{x}_{sat}}
\overset{\text{Patchify}}{\longmapsto}
\underset{   \in \mathbb{R}^{h' \times w' \times  T \times  (c*p_w*p_h)}}{\bm{x}_{sat}^{patched}}
\overset{\mathcal{L}^{c*p_w*p_h \to d}}{\longmapsto}
\underset{   \in \mathbb{R}^{h' \times w' \times  T \times  d}}{\bm{x}_{sat}^{proj}}
\end{equation*}
where, writing $h= h' *p_h$, $w=w' *p_w$ and $p_h$, $p_w \in \mathbb{N}$ are the pixel patch high and width, and $d$ is the embedding dimension. We identify again $\bm{x}_{sat}^{proj}$ with $\bm{x}_{sat}$.  The tensors $\bm{x}_{sat}$ and  $\bm{x}_{ts}$ are then transposed to be in  $\mathbb{R}^{ T \times  (h'*w') \times  d}$ and $\mathbb{R}^{T \times N_t \times d}$ , respectively, and are fed to the cross-attention layer. The operations of the the cross-attention layer are :
\begin{align}
\bm{z}_0 & = \text{LayerNorm}(\bm{x}_{ts}),\nonumber\\
\bm{z}_1 & = \text{LayerNorm}(\bm{x}_{sat}), \nonumber\\
\bm{z}_2 &=   \bm{x}_{ts} +  \text{MMSA}_{\text{RoPE}}(\bm{z}_0, \bm{z}_1,\bm{z}_1),  \nonumber\\
\bm{x}_{ts} &= \text{MLP}(\text{LayerNorm}(\bm{z}_2)) + \bm{z}_2. 
\end{align}

The Rotary Positional Encoding (RoPE)  introduced in the multi-head attention layer follows  \cite{su2024roformer, Boussif2023} and modifies the MSA definition from Eq. \eqref{MSA_eq} by introducing a rotation after the linear head embedding. The tensors $\bm{z}_0$ and $\bm{z}_1$ appearing in the first two arguments (query and key) of the MSA  undergo a rotational transformation along the feature dimension. The rotation angles are determined by the pixel and nodes position in longitude and latitude (see \cite{su2024roformer}, Eqs. (14)-(16) for details). The resulting vectors are subsequently used in the tensor contraction in Eqs.~\eqref{scaled_dot_prod} - \eqref{alph} between $\bm Q$ and $\bm K$. This rotation matrix ensures that part of the dot product captures the spatial relationships, such as node-to-node and node-to-pixel distances \cite{su2024roformer}.

\begin{center}
\begin{figure*}[ht]
    \centering
    \includegraphics[scale=0.45, trim = 10 5 5 7, clip]{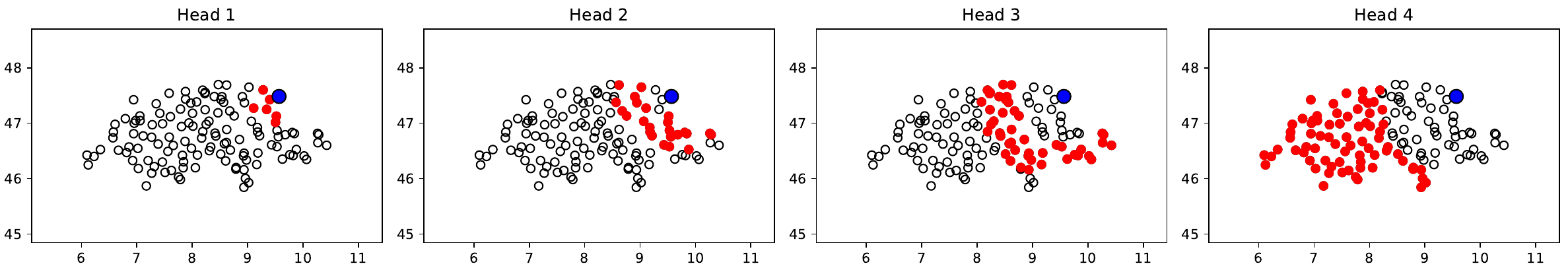} 
    \caption{Local attention heads for the node cross attention mechanism (4 heads). The blue dot is the central node ($i$-th node) and the red dots are the nodes where the mask $\bm{M}_{i.a}$ is set to 1. A similar local attention is used for the cross attention with the satellite images.}
    \label{local_attn}
\end{figure*}
\end{center}

Finally, we introduce a masked multihead scaled dot product attention layer (MMSA) that adds a masking term in the dot product of Eq. ~\eqref{alph}, that depends on the head number and the distance between nodes to capture local patterns. If the model has $n_h$ heads for the pixels-nodes cross-attention,  we create a masking tensor $\bm{M}$ of size $(N_t,h'*w',n_h)$. Each component $\bm{M}_{i.a} $ along the first and last axis is of size $h'*w'$ and is $0$ for pixels inside a ring of interior radius $r_a$ and outer radius $R_a$ centered on node $i$ and 1 otherwise, with  $R_a>r_a$, $r_1=0$, and $R_{n_h} = \infty$. The same is used for node-nodes cross attention, see Figure \ref{local_attn}.  Eq. \eqref{scaled_dot_prod} is then modified, for each head $a=1,...,n_h$, as
\begin{equation}
\label{alpha_local}
\alpha^{\text{local}}_{ \dots ij}(\bm{Q}, \bm{K}) := \frac{ \exp\left( \frac{\bm Q_{\dots i k} \bm  K_{\dots j k}}{\sqrt{d}} - \delta \bm{M}_{ija} \right) }{ \sum_{j'} \exp\left( \frac{\bm Q_{\dots i k} \bm  K_{\dots j' k}}{\sqrt{d}}  - \delta \bm{M}_{ij'a} \right)  }.
\end{equation}
where $\delta$ is a large positive real number.

\color{black}

\subsection{Models parameters and training strategies}
We used the same set of ground parameters for the SolarCrossFormer with and without satellite images. The image input size is  $96 \times 96 $ pixels, with a patch size of 4 to ensure high resolution inputs. 
We used a dynamic masking strategy for training. Not only were image pixels randomly masked, but we also randomly masked the whole observation sequence for a certain percentage of nodes, for each training batch.  We typically chose to have 10-16 randomly selected nodes per training batch and mask the past data of 15 \% of the nodes. The main model parameters are summarized in Table \ref{tab:parameters}.
 
\begin{table}[!htb]
      \caption{Main model parameters}
      \centering
\begin{tabular}{ll}
        \toprule
        \textbf{Parameter} & \textbf{Value} \\
        \midrule
        Number of MLP Heads & 1  / 3  \\
        Sat img. masking Ratio & 0.95 \\
        Time-Series Masking Ratio & 0.15 \\
        Embedding dim. & 128 \\
        Transformers depth & 3 \\
        Transformers heads & 4 \\
        MLP Ratio for heads & 3 \\
        Dimension per  head & 64 \\
        Dropout rate & 0.3 \\
        Decoder dim. & 64 \\
        Decoder depth & 3 \\
        Decoder heads & 4 \\
        Decoder dim. per head & 64 \\
        Decoder Input dim. per head  & 1 \\
        \bottomrule
    \end{tabular}
       \label{tab:parameters}
\end{table}

Masking was done after the linear embedding layers and embedding values were replaced by a learnable scalar. The aim of this masking was to use past data from other nodes and satellite images to infer the future GHI values of the masked nodes. Gradient  accumulation was used to fit the model gradients in the GPU memory.

To train the model we used the mean squared error (MSE) loss to get deterministic forecasts and the pinball loss function to learn multi-quantile predictions \cite{Koenker2001}. We chose to forecast the $[0.05, 0.5, 0.95]$ quantiles for each forecast point and use the 0.5 quantile (median) as the expected value and the 0.05 and 0.95 quantiles as confidence intervals.

\section{Experimental Setup}
\label{section_3}
\subsection{Datasets}
\begin{center}
\begin{figure}[t]
    \centering
    \includegraphics[scale=0.6, trim = 160 40 30 40, clip]{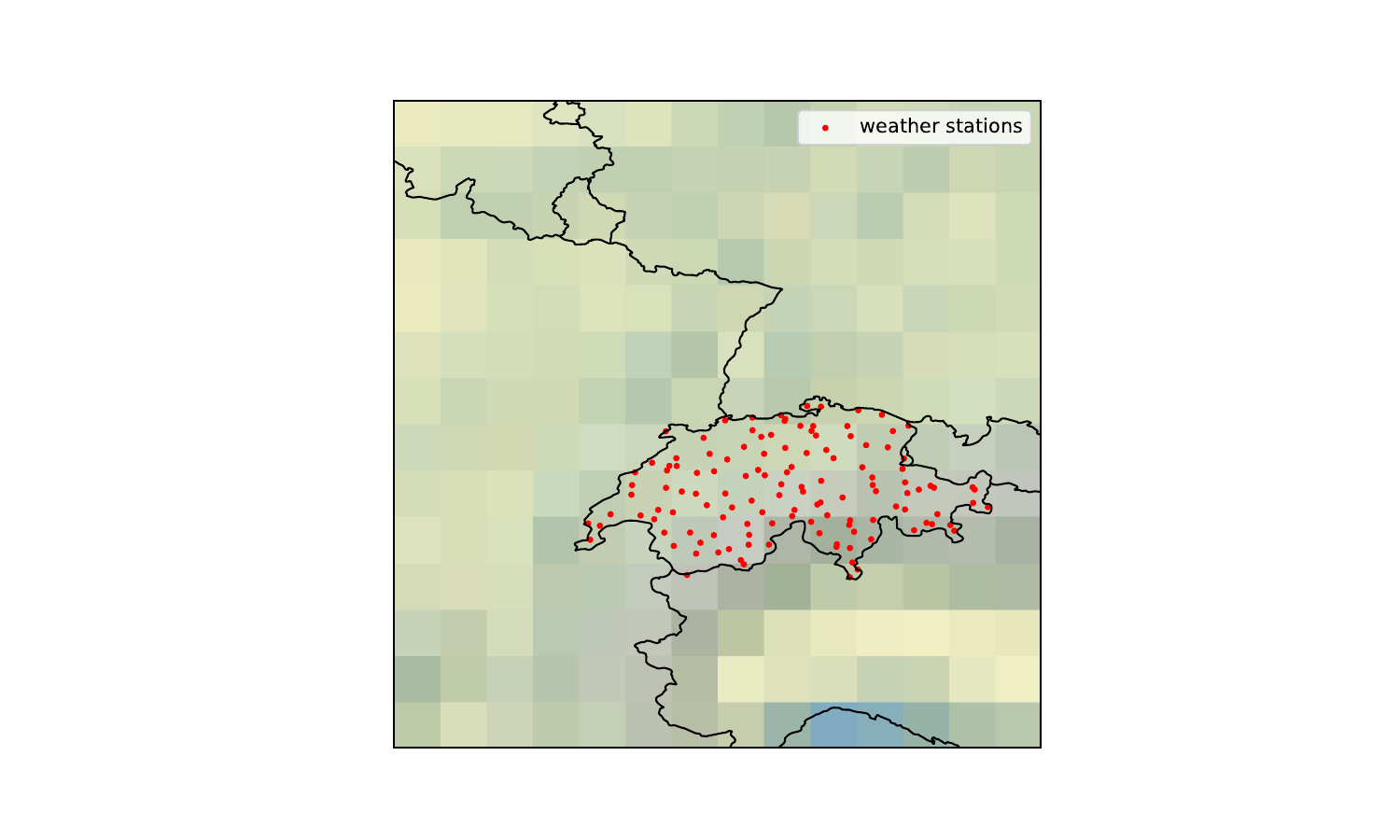} 
    \caption{Spatial distribution of the weather stations that conform the network of sensors. The area of the whole image corresponds to the area covered by the cropped satellite images. }
     \label{image_stations}
\end{figure}
\end{center}

We used two types of datasets in our study: time series data from weather stations and satellite imaging data. 
Time series of GHI, DNI, DHI, outside temperature, wind speed and direction, pressure and relative humidity, were obtained from the MeteoSwiss automatic measurement network\footnote{\url{https://gate.meteoswiss.ch/idaweb/more.do}}. We selected 127 weather stations whose locations spread across all Switzerland and provide the required measurements, see Figure \ref{image_stations}. The measurements have an original temporal resolution of 10 minutes but have been downsampled to 15 minutes to align with the desired temporal resolution. We also utilized data from satellite images (visual and infrared channels) from central Europe with a spatial resolution of 3km and a temporal resolution of 15 minutes from EUMETSAT MSG-4 satellite\footnote{\url{https://user.eumetsat.int/resources/user-guides/eumetsat-data-access-client-eumdac-guide}}. The channels used were: IR039, IR087, IR108 and VIS006. The original images were selected in a longitude-latitude bounding box $(-2.2, 35.2, 18.2, 55.6)$ with a size of 208$\times$ 208. However, they were cropped to a size of 96$\times$96 and centered around Switzerland to fit the GPU memory. Both datasets encompassed data from a 9-year period (2016-2024). Data from 2016 to 2023 was used for training while the 2024 subset was used for evaluation. Both datasets can be downloaded for research purposes.

\subsection{Performance Metrics}
The performance of the SolarCrossFormer model was evaluated using several metrics for both deterministic and probabilistic forecasts. 

The peak normalized root mean-squared error (NRMSE) and the peak normalized mean absolute error (NMAE) were used as metrics for deterministic forecasts. They are defined, for site $v$ and lead time $i$ as:
\begin{equation*}\label{eq:nrmse}
NRMSE(v, i) = \sqrt{\frac{1}{|\mathcal{T}|} \sum_{t \in \mathcal{T}} \left(\frac{\hat y_{v}(t+i)-y_{v}(t+i)}{{y^{max}}}\right)^2},
\end{equation*}
\begin{equation*}\label{eq:nmae}
NMAE(v, i)=  \frac{1}{|\mathcal{T}|}\sum_{t \in \mathcal{T}} \frac{|\hat y_{v}(t+i)- y_{v}(t+i)|  }{y^{max}}, 
\end{equation*}
where $y_v(t)$ and  $\hat{y}_v(t)$ denote the ground truth GHI and the predicted GHI, respectively, of site $v$ at time $t$. The maximum GHI  $y^{max}$ is observed over the evaluation period $\mathcal{T}$ and set to $1.3\text{ }kWm^{-2}$, while $|\mathcal{T}|$ is the number of time steps in the evaluation interval $\mathcal{T}$. Night time can be excluded from the computation (by excluding points where $y_v(t)=0$) and this will be clarified in each case.  An additional metric we report is the mean absolute percentage error (MAPE), that is defined as:
\begin{equation*}\label{eq:mape}
MAPE(v, i)=  \frac{1}{|\mathcal{T}|}\sum_{t \in \mathcal{T}} \frac{|\hat y_{v}(t+i)- y_{v}(t+i)|  }{ y_{v}(t+i)}.
\end{equation*}
To avoid giving a large weight in the MAPE computation to tails of the daylight, we only included points in the MAPE when the observed irradiance is higher than $100 W/m^2$.

To evaluate the reliability, sharpness and resolution of the probabilistic forecasts we used the prediction interval coverage probability (PICP), the prediction interval average width (PINAW) and the normalized continuous rank probability score (NCRPS) metrics. They are defined in the following equations for site $v$ and lead time $i$.
\begin{equation*}\label{eq:picp}
PICP(v, i)=  \frac{1}{|\mathcal{T}|}\sum_{t \in \mathcal{T}} \mathbb{\chi}(y_{v}(t+i)\in \hat{PI}(t+i)), 
\end{equation*}
\begin{equation*}\label{eq:pinaw}
PINAW(v, i)=  \frac{\sum_{t \in \mathcal{T}} (\hat{y}_{v}^{\beta}(t+i)-\hat{y}_{v}^{\alpha}(t+i))}{\sum_{t \in \mathcal{T}}y_v(t+i)}, 
\end{equation*}
where $\hat{PI}(t+i)=[\hat{y}_{v}^{\alpha}(t+i),\hat{y}_{v}^{\beta}(t+i)]$ denotes the prediction interval between the $\alpha=0.05$ and $\beta=0.95$ quantiles and $\mathbb{\chi}(\cdot)$ denotes the indicator function whose value is 1 if its argument is true, or zero otherwise. The NCRPS is defined as
\begin{align*}\label{eq:crps}
&NCRPS(v, i)=  \\
&\frac{1}{|\mathcal{T}|y^{max}}\sum_{t \in \mathcal{T}} \int_{-\infty}^{\infty}\left[ \hat{F}_{v}(x,t+i) - \mathbb{\chi}(x\geq y_{v}(t+i))\right]^2 dx, 
\end{align*}
where $\hat{F}_{v}(x,t+i)$ denotes the cumulative predictive distribution at site $v$ and lead time $t+i$. Target value for the PICP is 0.9 (or 90\%) since the prediction interval of the selected quantiles is 90\%. Regarding the NCRPS, the closer to zero the metric is the better.

\subsection{Benchmark Models}

We conducted a thorough comparison of SolarCrossFormer with state-of-the-art models tailored for day-ahead solar irradiance forecasting. The first benchmark model is the graph-convolutional long-short term memory (GCLSTM) developed in \cite{Simeunovic2022a} for multi-site PV forecasting and adapted for day-ahead multi-site irradiance forecasting in \cite{Carrillo2023}. This model uses the data from the MeteoSwiss network of weather stations to forecast the GHI for the same locations as its input network. The second benchmark model is the CrossVivit model from Boussif \textit{et. al.} \cite{Boussif2023}, that forecasts the day-ahead GHI for a single site using the past data from the site and the spatial context from the satellite imaging data. We made some changes to the CrossVivit architecture presented in \cite{Boussif2023} to improve  bottlenecks and overfitting issues encountered when training it: the depth and number of heads were reduced; in the final MLP layers we replaced the last activation function of the authors (a ReLu) by a CELU activation, followed by a linear mapping, to overcome dead neurons not training during a large number of gradient steps. We used the same hidden dimensions and parameters for the transformers in CrossVivit as for the SolarCrossFormer; see  Table \ref{tab:parameters}.  Moreover, we also added the possibility to input the future clear sky irradiance values in the decoder. We used the same data as in SolarCrossFormer, \textit{i.e.}, satellite imaging data and weather measurement data from all locations of the MeteoSwiss network, to train the CrossVivit model. However, we evaluated one site at a time. The third benchmark model is the SolarFusionNet from Jing \textit{et. al.} \cite{Jing2024}. We adapted the authors's implementation to our experimental setting by training it on our dataset, sampling nodes at random from the available set for each training batch and forecasting the next 24 hours. The fourth benchmark was a commercial NWP solution that yields GHI forecasts for 24 hours ahead with hourly temporal resolution. We also compared the SolarCrossFormer model with a modified version of itself that doesn't use satellite data. We report this model as SolarCrossFormer (no img.). Each model type (SolarCrossFormer, CrossVivit, SolarFusionNet, GCLSTM, and SolarCrossFormer without images) was trained with five different random seeds for each loss function. The best-performing models for each type were selected based on the evaluation year 2024, taking the best performing models (\text{i.e.}, minimizing the training loss function) both over the number of gradient steps and seed variations.

\section{Experimental Results}
\label{section_4}
\subsection{Multi-site Irradiance Forecasting}
\begin{center}
\begin{figure}[t]
    \centering
    \includegraphics[scale=0.3, trim = 40 40 40 80, clip]{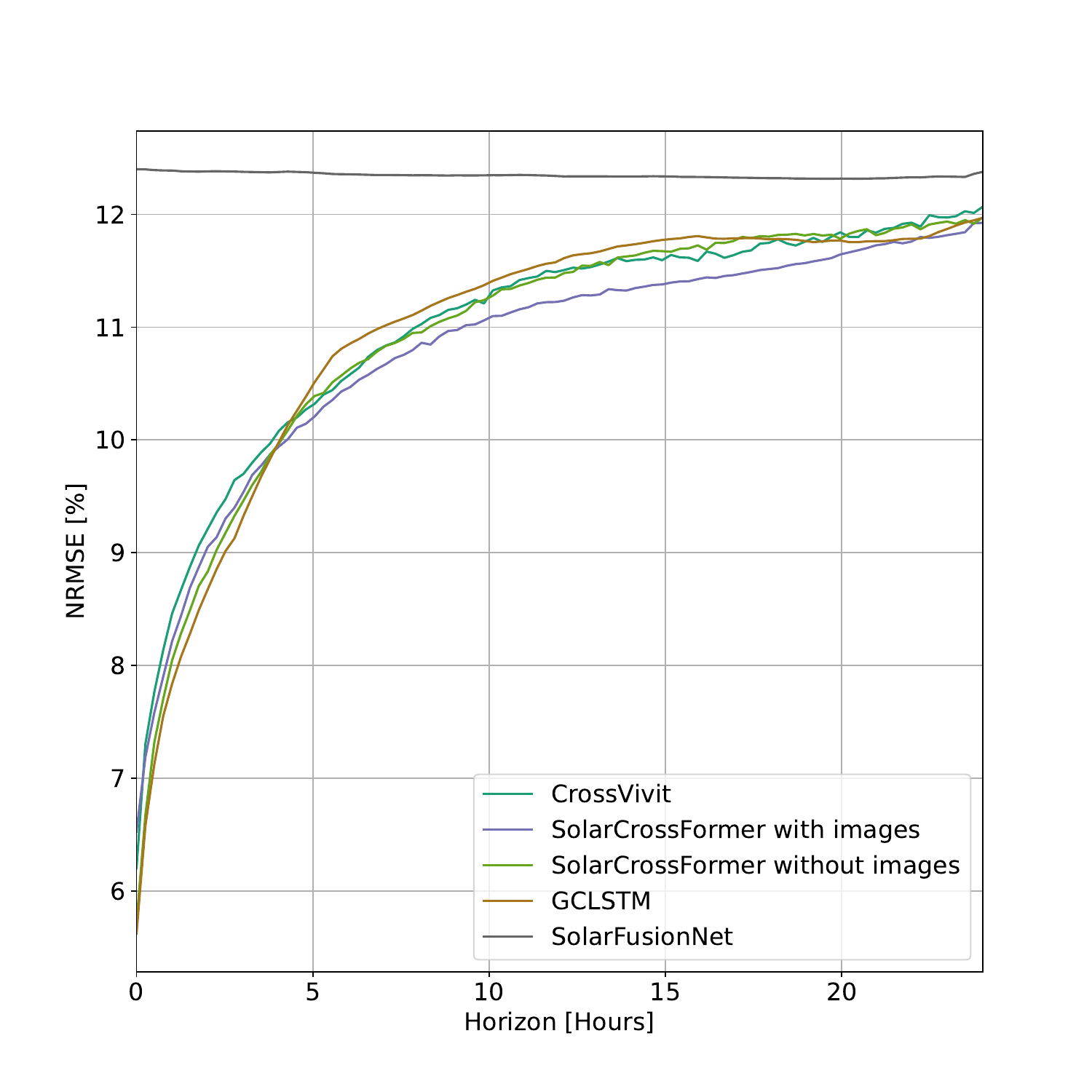} 
    \caption{Errors over the horizon: NRMSE for the best models trained under the MSE loss.}
    \label{mse_nrmse}
\end{figure}
\end{center}

We begin by presenting the overall accuracy results for the evaluation year 2024, based on GHI records from 127 ground stations in Switzerland, obtained from the MeteoSwiss network. Figure \ref{mse_nrmse} shows the NRMSE over the 24 hour prediction horizon (in steps of 15 minutes) for the best model of each type trained with the MSE loss. The solid line is the median NRMSE across the 127 nodes. Forecasts for night-time hours were excluded from the computation. As can be see on Figure \ref{mse_nrmse}, the performance of the best models of each type are close except for the SolarFusionNet model. However, the SolarCrossFormer with satellite images clearly outperforms the other four models for time horizons between 5 and 24 hours. The GCLSTM model achieved the best median NRMSE for the first 4 hours due to its recursive architecture that takes advantage of the most recent measurements from the past sequence. Over the five seeds results, we consistently found that the SolarCrossFormer model achieved lower error than the other models globally and over most of the horizon. One explanation for the poor performance of the SolarFusionNet model is that the model takes node and image coordinates as inputs to the forward pass, but these are not explicitly exploited. The attention mechanism is applied only in the temporal dimension, while spatial information from the images is propagated through 2D convolutions in the LSTMs. Unlike architectures such as CrossViVit or the proposed SolarCrossformer, no cross-attention with RoPE is used. Thus, the forward pass is essentially coordinate-agnostic. While this may be suitable for single-site training, it appears less effective when training across multiple sites simultaneously. We therefore decided to exclude the SolarFusionNet from the rest of the experiments.

\begin{center}
\begin{figure}[t]
    \centering
    \includegraphics[scale=0.3, trim = 5 40 40 80, clip]{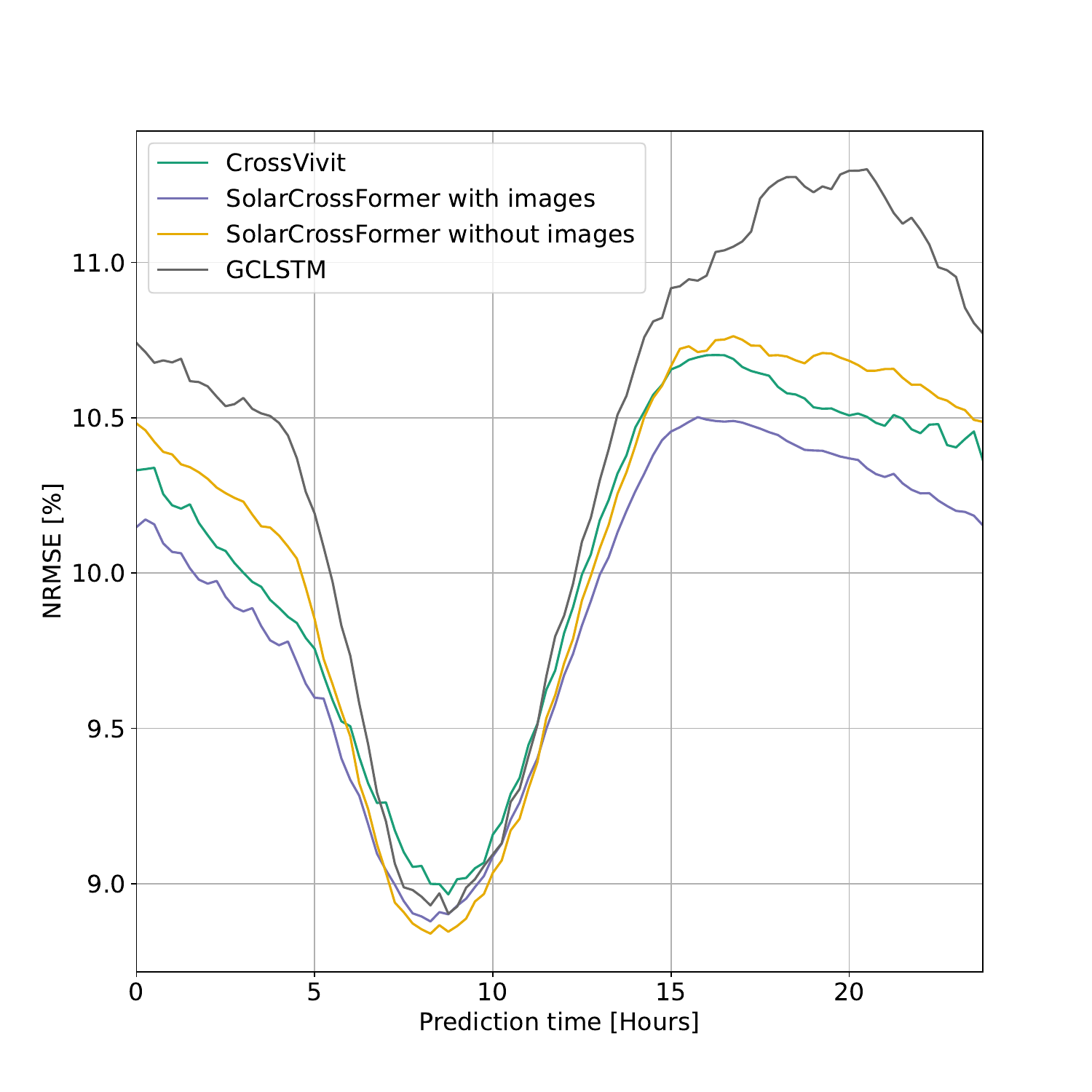} 
    \caption{Errors by prediction time: NRMSE for the best models trained under the MSE loss.}
    \label{pred_nrmse}
\end{figure}
\end{center}

\begin{center}
\begin{figure*}[t]
    \centering
    \includegraphics[scale=0.35, trim = 5 40 40 80, clip]{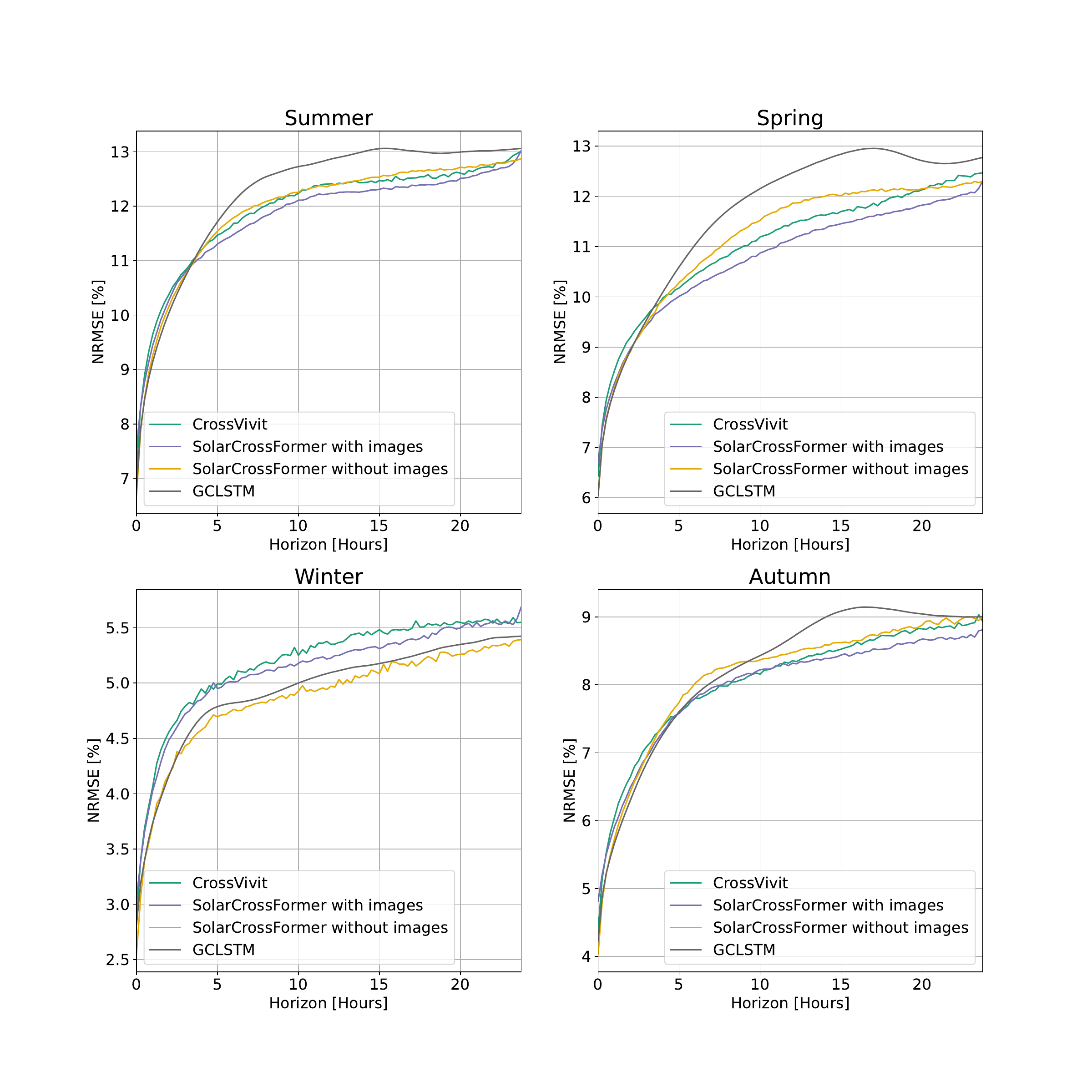} 
    \caption{Errors over the horizon: NRMSE by seasons for the best models trained under the MSE loss.}
    \label{season_nrmse}
\end{figure*}
\end{center}

Horizon-averaged NRMSE,  NMAE and MAPE for this set of models are given on Table \ref{tab:mse_trained}, both when forecasts for night-time hours have been excluded from the computation or included in the computation (except for MAPE to avoid division by zero). Best results are outlined in bold. When averaging over the horizon, the SolarCrossFormer model with satellite images outperforms the second best model by 0.17\% on NRMSE (and by 0.8 \% on MAPE), which amounts to $2.21 Wm^{-2}$ when rescaling with the normalization factor. However, on NMAE, CrossVivit outperforms SolarCrossFormer by 0.04\% ($0.52 Wm^{-2}$). 

Horizon-averaged NRMSE, NMAE and MAPE for models trained with the pinball loss are given on Table \ref{tab:quantiles_trained}, both when forecasts for night-time hours have been excluded from or included in the computation. Best results are outlined in bold. When averaging over the horizon, the best SolarCrossFormer model trained with satellite images outperforms the second best model by 0.3\% on NRMSE, and by 0.07 \% on NMAE (notice that optimizing the pinball loss for the median amounts to minimizing the MAE).

\begin{table}[htbp]
   \caption{Mean accuracy over the 24h horizon in 2024. All models were trained with the MSE loss.} 
   \label{tab:mse_trained}
   \footnotesize
    \centering
    	\begin{tabular}{l|c|c|c|c}
	\toprule
	Model & NRMSE & NMAE &MAPE \\
	 & [\%] & [\%]  & [\%]\\
	\midrule
	GCLSTM  & 10.58 & 6.34 & 44.05\\
	CrossVivit &10.32 & \bf{6.32} & 43.76  \\
	SolarFusionNet & 11.67& 7.70& 52.96 \\
	SolarCrossFormer (no img.) &10.36 & 6.37& 43.49\\
	SolarCrossFormer &\bf{10.15} & 6.36 & \bf{42.70} \\
	\midrule
         \textit{with night} \\
	\midrule
	GCLSTM & 8.75 & 4.24 & -\\
	CrossVivit &8.49 &\bf{4.22}& -\\
	SolarFusionNet & 9.61& 5.10& -  \\
	SolarCrossFormer (no img.)& 8.52 & 4.29 &- \\
	SolarCrossFormer& \bf{8.36} & 4.39& - \\
	\bottomrule
	\end{tabular}
\end{table}

\begin{table}[htbp]
   \caption{Mean accuracy over the 24h horizon in 2024.  The models  were trained with the pinball loss.} 
   \label{tab:quantiles_trained}
   \footnotesize
    \centering
    	\begin{tabular}{l|c|c|c|c}
	\toprule
	Model  & NRMSE & NMAE &MAPE \\
	 & [\%] & [\%] & [\%]\\
	\midrule
	GCLSTM & 10.97 & 6.18 &43.42\\
	CrossVivit &10.95 & 6.31& 45.31\\
	SolarCrossFormer (no img.)  &10.89 & 6.16 & \bf{42.67}\\
	SolarCrossFormer &\bf{10.59} & \bf{6.09}&43.55 \\
	\midrule
         \textit{with night}\\
	\midrule
	GCLSTM & 9.06 & 4.09 \\
	CrossVivit  & 9.00 & 4.21 & -\\
	SolarCrossFormer (no img.)  &8.95 & 4.07 & - \\
	SolarCrossFormer  &\bf{8.71} & \bf{4.05}& -  \\
	\bottomrule
	\end{tabular}
\end{table}

Figure \ref{pred_nrmse} shows the average NRMSE of the models by prediction time (i.e. averaging over the nodes and the horizon). The integration of imagery data and data from the network of sensors consistently enhances the model performance, particularly during night time or late-day forecasts over the 24-hour horizon. CrossVivit and the SolarCrossFormer clearly outperform the other models at these predictions period, the SolarCrossFormer delivering the best results. We also analysed the forecasting error by comparing the performance on different seasons. Figure \ref{season_nrmse} shows the NRMSE over the 24 hours horizon for the 4 seasons of 2024. We observe a clear advantage in spring, summer and autumn seasons for the SolarCrossformer model. Visualizations of the forecasted trajectories at different moment of the day, comparing  GCRNN, CrossViVit and the SolarCrossFormer, are given in the appendix, Figures \ref{mod_comp_1} to \ref{mod_comp_3}. 

The probabilistic metrics for models trained using the pinball loss function are reported in Table \ref{tab:quantiles_trained_proba}. For the PICP, GCLSTM achieved the best result (interval coverage is the closest to 90\%), with SolarCrossFormer without images coming in second place with 91.4 \%. The models that use satellite images as input had a larger coverage though still close to the desired 90\%. For the PINAW, the models that only use data from the network of sensors, SolarCrossFormer and GCLSTM, had the narrowest interval width. For the NCRPS, GCLSTM achieved the lowest score though SolarCrossFormer remain close in performance (0.01\% difference).

\begin{table}[htbp]
   \caption{Probabilistic metrics for the models trained with the pinball loss. Forecasts at night are excluded from the calculation.} 
   \label{tab:quantiles_trained_proba}
   \footnotesize
    \centering
    	\begin{tabular}{l|c|c|c}
	\toprule
	Model & PICP [\%] & PINAW [\%] & NCRPS [\%]   \\
	\midrule
	GCLSTM &\bf{90.76} & 132.50 &  \bf{2.96}\\
	CrossVivit &  93.97 & 140.14 & 3.07 \\
	SolarCrossFormer  & 93.69 & 143.02 &  2.97 \\
	SolarCrossFormer (no img.)  & 91.40 & \bf{130.07} & 2.98 \\
	\bottomrule
	\end{tabular}
\end{table}

Summarizing the findings from Figure \ref{mse_nrmse} to Table \ref{tab:quantiles_trained_proba}, we observe that the SolarCrossFormer trained with satellite images consistently outperforms CrossVivit and models trained without images. However, the improvement margin is relatively small. This suggests that the sequence of satellite images provides only limited additional information for predicting GHI trends at the studied sites, as compared to using solely information from other nodes. Notably, CrossVivit, despite incorporating a vision transformer to extract features from satellite images in its initial layers, does not outperform the shallower and simpler architecture SolarCrossFormer. Adding a vision transformer layer to the SolarCrossFormer did not yield accuracy improvements either, nor did incorporating factorized time and spatial attention in the vision transformer or in the CrossFormer layers (see \textit{e.g.} \cite{arnab2021vivit}). Although the exact reason is uncertain, we attribute this lack of improvement mainly to two factors. First, adding more layers—especially vision transformer layers for high-resolution images—appears to increase the model’s tendency to overfit. Second, rather than focusing on extracting cloud dynamics, the vision transformer seems to primarily memorize images instead of extrapolating cloud motion. This limitation might potentially be mitigated by incorporating an additional forecasting term on future images into the loss function.

Furthermore, we identified an important aspect that was not thoroughly discussed in \cite{Boussif2023} but proved to be highly influential: models trained with satellite images, particularly CrossVivit, tend to overfit significantly. To mitigate this issue, it was necessary to implement various overfitting reduction strategies, including input masking, dropout, cosine annealing with warm restarts, and layer size reduction. Early stopping was  essential, as the networks quickly memorized the eight years of training data and tended to predict what was seen in the past instead of learning the ``physics'' of cloud propagation. In fact, the present study employs shallower CrossVivit models compared to those in \cite{Boussif2023}, which used a training dataset of similar size and likely faced severe overfitting challenges. The authors in \cite{Boussif2023} had best results when using a high random masking rate of the images for training (masking rate randomly chosen between 0  and 0.99). We have confirmed that the random masking is a major factor for avoiding strong overfitting. This masking might however reduce the amount of  local information from the satellite images that the model uses to forecast the GHI. Thus, it might be reducing the final performance gain that could be expected from adding high resolution satellite image information.

\subsection{Comparison with NWP-based forecasts}
\begin{center}
\begin{figure}[t]
    \centering
    \includegraphics[scale=0.31, trim = 35 40 40 80, clip]{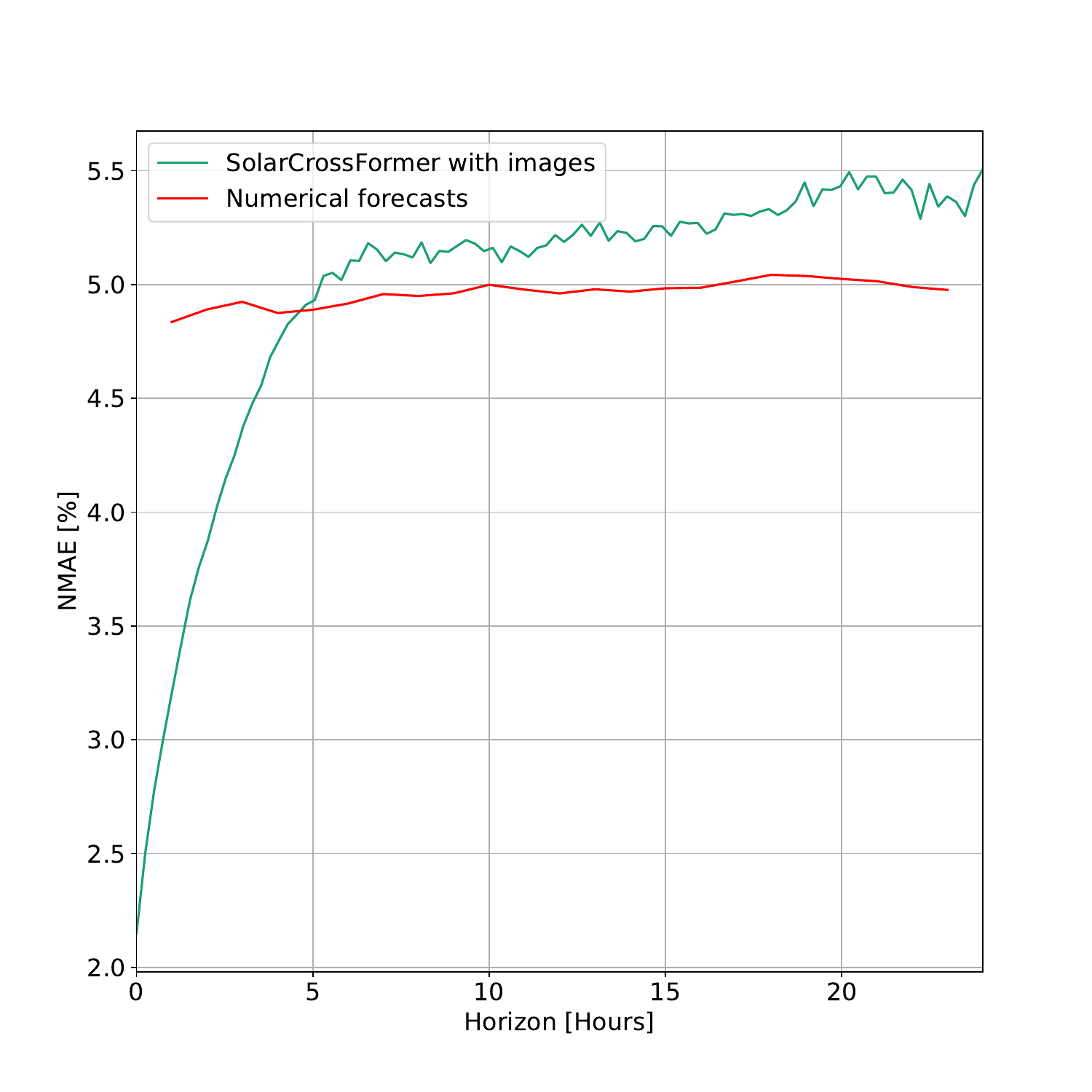} 
    \caption{Error over the horizon: Comparison between a commercial  NWP irradiance forecast and the best SolarCrossFormer trained with the pinball loss, from 01.10.2024 to 31.12.2024, in Neuchâtel.}
     \label{meteo_nmae}
\end{figure}
\end{center}

\begin{figure}[t]  
   \begin{center}
    \includegraphics[scale=0.31, trim = 35 40 40 80, clip]{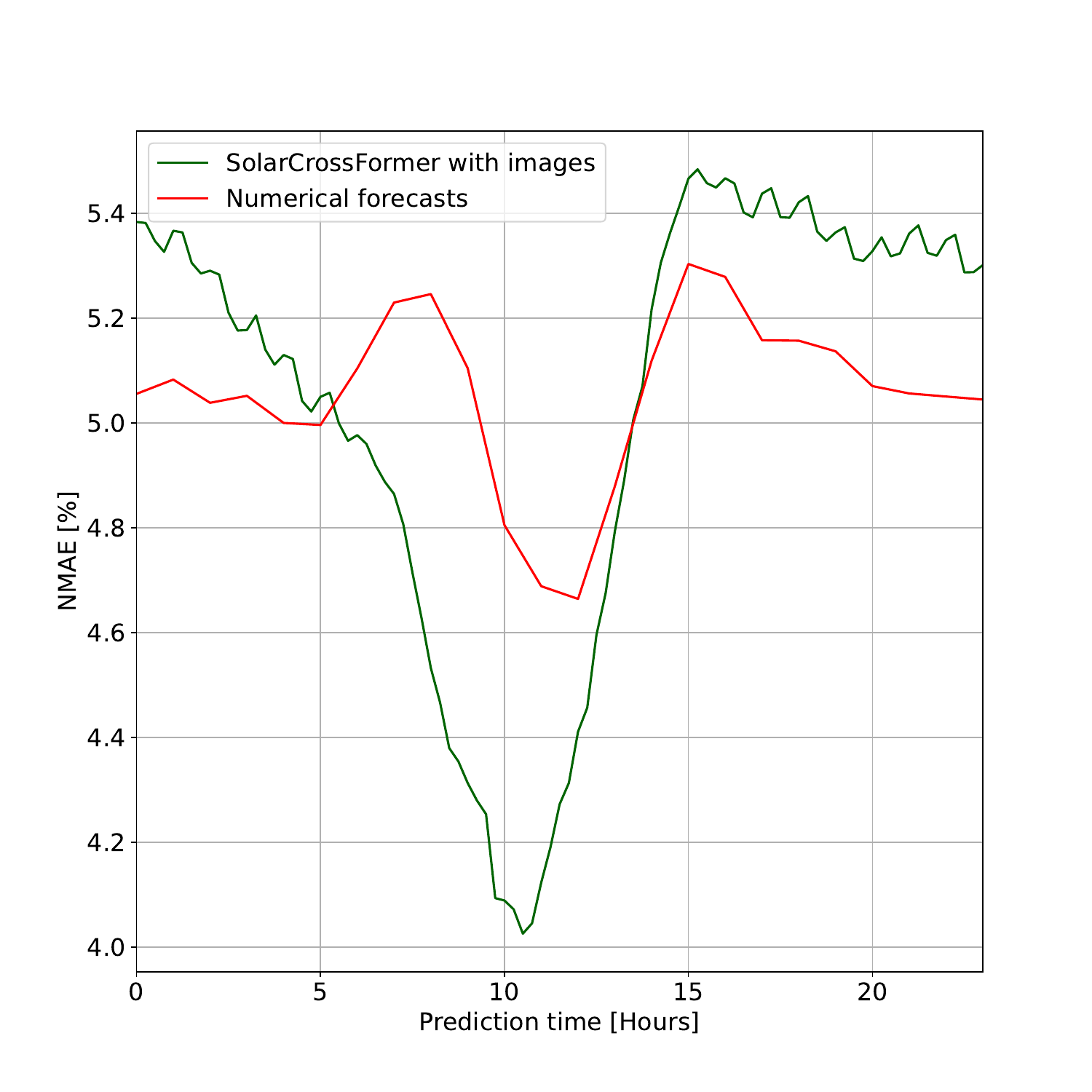} 
    \caption{Error by prediction time: Comparison between a commercial  NWP irradiance forecast and the best SolarCrossFormer trained with the pinball loss, from 01.10.2024 to 31.12.2024, in Neuchâtel}
     \label{nwp_vs_solarformer_pred_time}
   \end{center}
\end{figure}

We evaluated the forecasts generated by the best-performing model trained using the pinball loss against GHI forecasts provided by a commercial weather service in Switzerland, which are based on numerical weather predictions (NWP). While the NWP data is updated every three hours, the commercial provider enhances its forecasts hourly by integrating ground-based measurements, resulting in both hourly resolution and update frequency. We compare the performance over three months in 2024, from beggining of Octuber to end of December, in Neuchâtel, Switzerland. The NMAE over the forecasting horizon is presented in Figure \ref{meteo_nmae}. Results show that SolarCrossFormer outperforms the NWP-based forecasts for horizons up to 5 hours ahead. In Figure \ref{nwp_vs_solarformer_pred_time}, we compare the NMAE between the NWP-based forecast and the SolarCrossFormer by prediction time, averaged over the 24h horizon. For prediction times from 5 a.m. to 2 p.m., the SolarCrossFormer has a lower NMAE than the NWP-based forecast. The gap in the prediction accuracy in favour of SolarCrossFormer is largest in the middle of the morning, where the latter model exhibits up to a 1\% improvement as compared to NWP. The superior performance of the NWP-based service during night time and late-day forecasts can be attributed to the limited availability of observational data, such as satellite imagery and ground-based measurements, during these periods. In contrast, NWP models rely on numerical simulations that remain robust regardless of time of day, providing a more comprehensive basis for forecasting over the next 24 hours. This behaviour is consistent with findings in the literature, which highlight that the choice of data sources and modeling approaches often depends on the forecasting horizon. Specifically, satellite and ground-based data yield higher accuracy for intra-day forecasts, whereas NWP-based models are more effective for day-ahead and longer-term predictions (see \cite{Yang2022,Antonanzas2016} for further discussion).

The accuracy results, averaged over the forecast horizon, are presented in Table \ref{tab:neuch_meteo}. This analysis compares the performance of SolarCrossFormer models trained with mean squared error (MSE) and pinball loss functions against forecasts derived from NWP-based methods, excluding nights. Both the NWP and SolarCrossFormer models exhibit comparable accuracy in terms of mean absolute error (MAE). Overall, the NWP-based forecasts demonstrate slightly superior performance across the full prediction horizon. However, it is important to highlight that NWP-based approaches typically involve substantial computational and storage requirements, which often lead to implementations with reduced temporal and spatial resolution. In contrast, deep learning models such as SolarCrossFormer can achieve computational speed-ups by a factor 100, offering a significant advantage for scalable and real-time forecasting applications \cite{Carrillo2022}. Additionally, the evaluation period, from October to December 2024, was particularly challenging due to frequent foggy conditions. Figure \ref{nwp_vs_solarformer_sunny} in the Appendix illustrates examples of forecast trajectories on a sunny day following a foggy day in Neuchâtel. 

\begin{table}[htbp]
   \caption{Model accuracy in Neuchâtel from 1.10.2024 to 31.12.2024.} 
   \label{tab:neuch_meteo}
   \footnotesize
    \centering
    	\begin{tabular}{l|c|c|c|c}
	\toprule
	Model  & training & NRMSE & NMAE &MAPE  \\
	 &loss&    [\%] &  [\%]  \\
	\midrule
	SolarCrossFormer  & MSE&  7.67 & 5.18 & 49.33  \\
	SolarCrossFormer   & Pinball & 7.90 & 4.97& 48.96  \\
	NWP &- & \bf{7.13} &  \bf{4.96} &  \bf{40.73}\\
	\bottomrule
	\end{tabular}
\end{table}

\subsection{Forecasting without ground stations data}
\begin{center}
\begin{figure}[ht]
    \centering
    \includegraphics[scale=0.5]{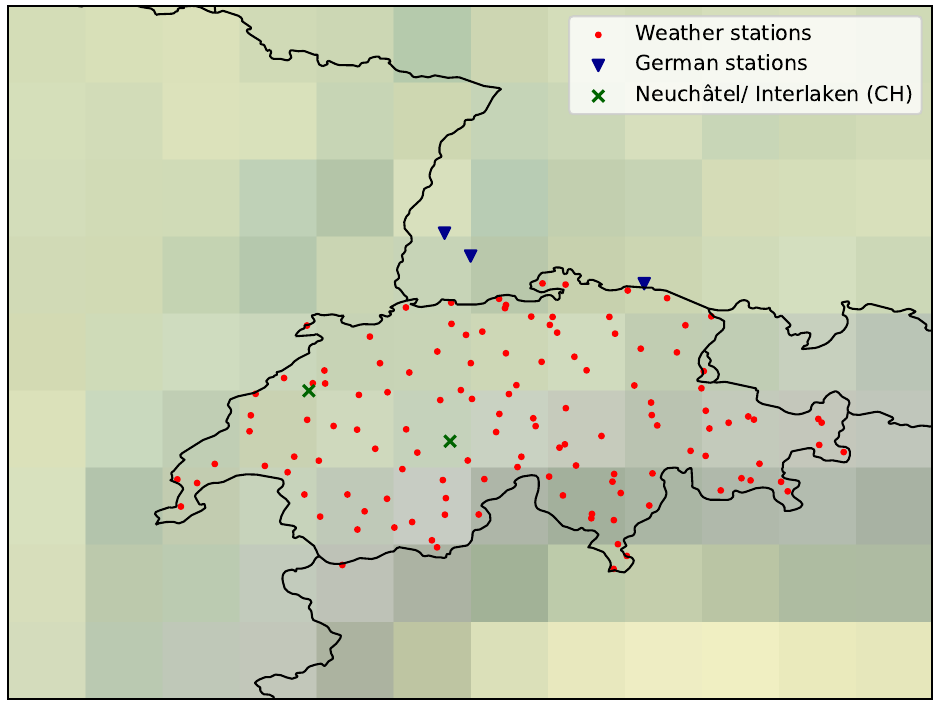} 
    \caption{Locations of the nodes for forecasting at ``unseen'' locations:  forecasts for triangle- and cross-marked nodes are evaluated without any past ground observations at these nodes.  }
    \label{nodes_map_rec}
\end{figure}
\end{center}

The SolarCrossFormer models were trained by randomly masking past data from a proportion of the nodes for each training batch. Therefore they are capable of forecasting GHI at locations where no local observations are available, using other nodes data and/or the satellite images. We tested the SolarCrossFormer models with and without image data to check the accuracy gain in situations where we forecasted the GHI at ``unseen'' nodes. For this purpose, we selected three meteorological stations in Germany outside the boundaries of the points used for training (as we trained only with Swiss automatic stations) and two stations that were included in the training data but were deliberately masked entirely during evaluation, see Figure \ref{nodes_map_rec}.

\begin{center}
\begin{figure*}[h]
    \centering
    \includegraphics[scale=0.45,trim = 40 40 30 40, clip]{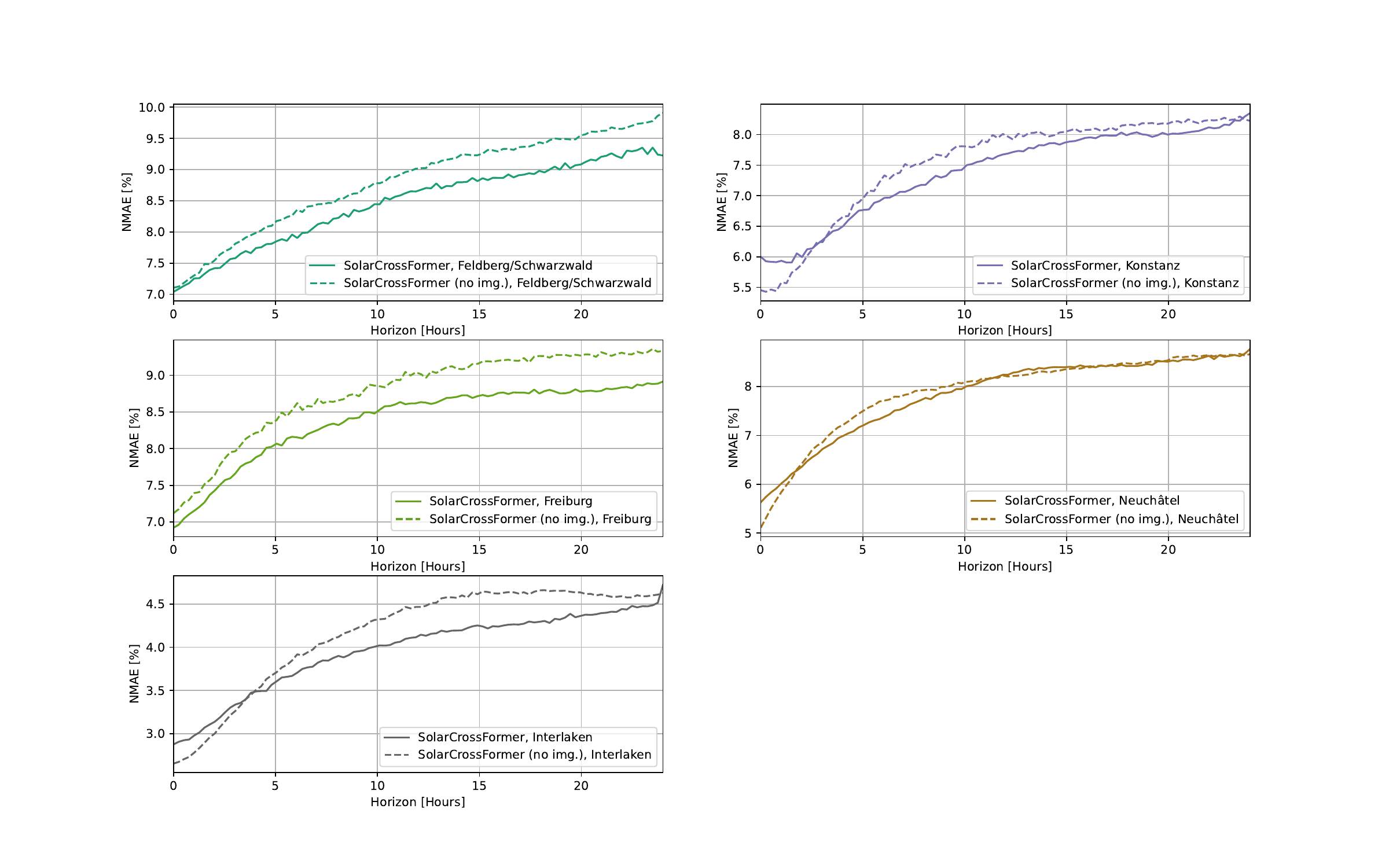} 
    \caption{Errors over the horizon: Comparison of the prediction error (NMAE) on five nodes over the year 2024, where past station data were entirely masked and no local measurement were available.}
    \label{rec_nmae_nodes}
\end{figure*}
\end{center}

Evaluation accuracy results are summarized in Table \ref{tab:table_NMAE_rec} for models trained with pinball loss, considering both when ground-measured past data was entirely masked or where it was fully available. The SolarCrossFormer incorporating satellite images consistently outperformed the model without satellite image inputs. This advantage was particularly evident at nodes in Germany, where the performance gap reached 0.5–0.6 \% in Freiburg and Konstanz in favor of the SolarCrossFormer with satellite images. These findings are further supported by the NMAE horizon plots for all five nodes in Figure \ref{rec_nmae_nodes}, showcasing the best-performing models trained with pinball loss.

\begin{center}
\begin{table}[htbp]
    \caption{Forecast accuracy for models trained on pinball loss for the year 2024.}
    \label{tab:table_NMAE_rec}
   \footnotesize
    \centering
    \begin{tabular}{l|c|c|c|c|c|c}
\toprule
Model &  Name & NRMSE  & NMAE &  CRPS \\
& &[\%] &[\%] & [\%]\\
\midrule
SolarCrossFormer &Feldberg  & \bf{12.74} & \bf{8.46} &  \bf{4.08} \\
SolarCrossFormer  & Konstanz & \bf{11.24}    & \bf{7.40} & \bf{3.59} \\
SolarCrossFormer  & Freiburg  & \bf{12.45}  & \bf{8.39} &  \bf{4.23} \\
SolarCrossFormer   & Neuchâtel  & \bf{12.24} & \bf{7.85} & \bf{3.74} \\
SolarCrossFormer  & Interlaken & \bf{8.63} & \bf{3.97} & \bf{1.94} \\
SolarCrossFormer (no img.) & Feldberg  & 13.11 & 8.81 &  4.39 \\
SolarCrossFormer (no img.) & Konstanz & 11.85  & 7.54 & 3.71 \\
SolarCrossFormer (no img.) & Freiburg  & 13.03  & 8.77 & 4.49 \\
SolarCrossFormer (no img.) & Neuchâtel & 12.31   & 7.90 & 3.76 \\
SolarCrossFormer (no img.) &Interlaken & 9.03  & 4.16 & 2.01 \\
\midrule
 \textit{without masking}\\
\midrule
SolarCrossFormer  & Neuchâtel & \bf{11.93}  & \bf{7.54} & \bf{3.62} \\
SolarCrossFormer   & Interlaken & \bf{8.53}  &\bf{3.88} & \bf{1.90} \\
SolarCrossFormer (no img.)  & Neuchâtel & 12.30 & 7.72 & 3.70 \\
SolarCrossFormer (no img.) & Interlaken & 9.07 &4.09 & 1.98 \\
\bottomrule
\end{tabular}
\end{table}
   \end{center}

\begin{center}
\begin{figure}[h]
    \centering
    \includegraphics[scale=0.26, trim = 25 0 10 0, clip]{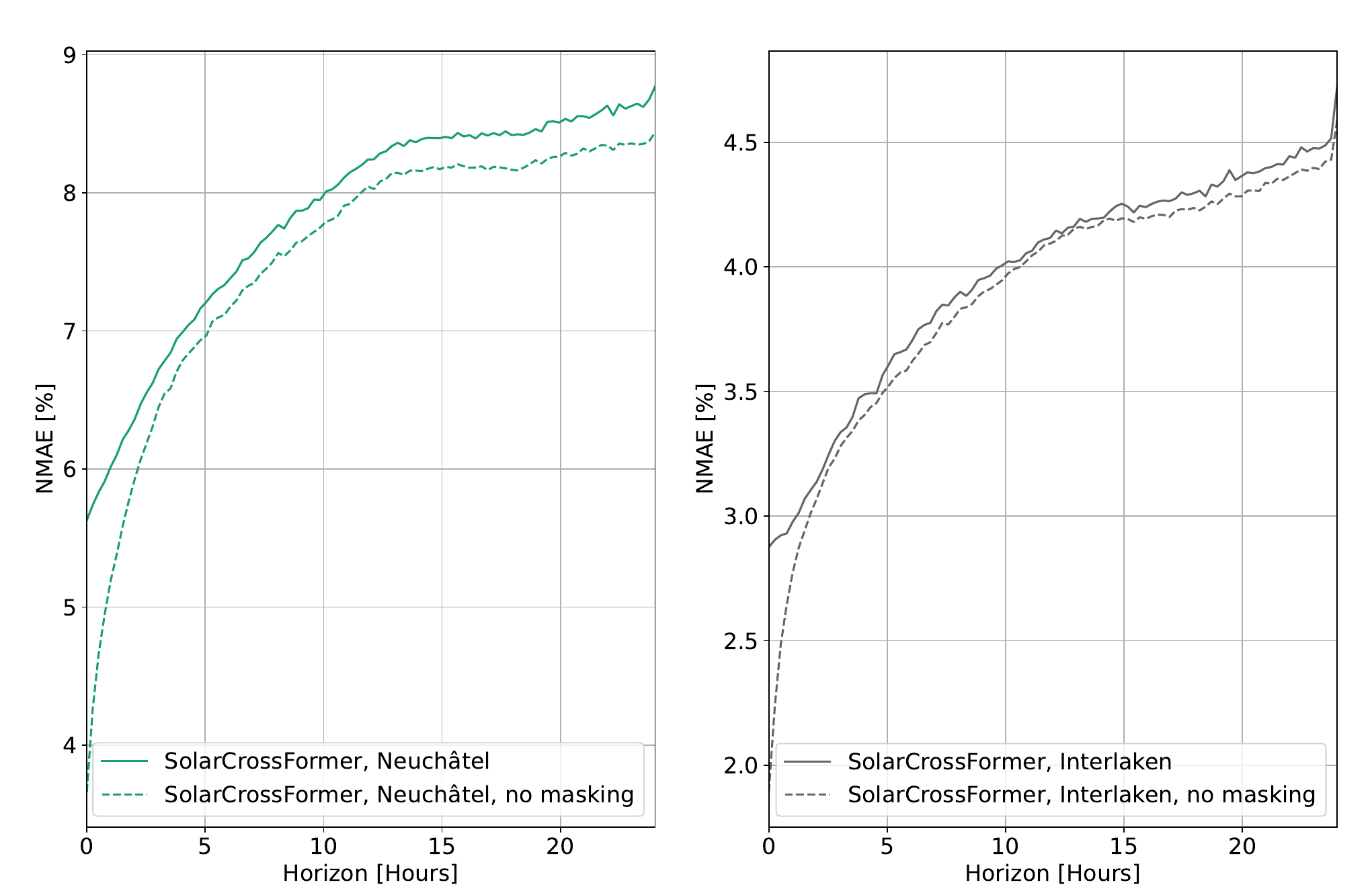} 
    \caption{Masked vs. non-masked node errors for the SolarCrossFormer in Neuchâtel and Interlaken, Switzerland, over the 24 hours horizon.}
    \label{comp_with_without_mask}
\end{figure}
\end{center}

\begin{flushleft}
\begin{figure}[h!]
    \includegraphics[width=8.7cm, height=5cm]{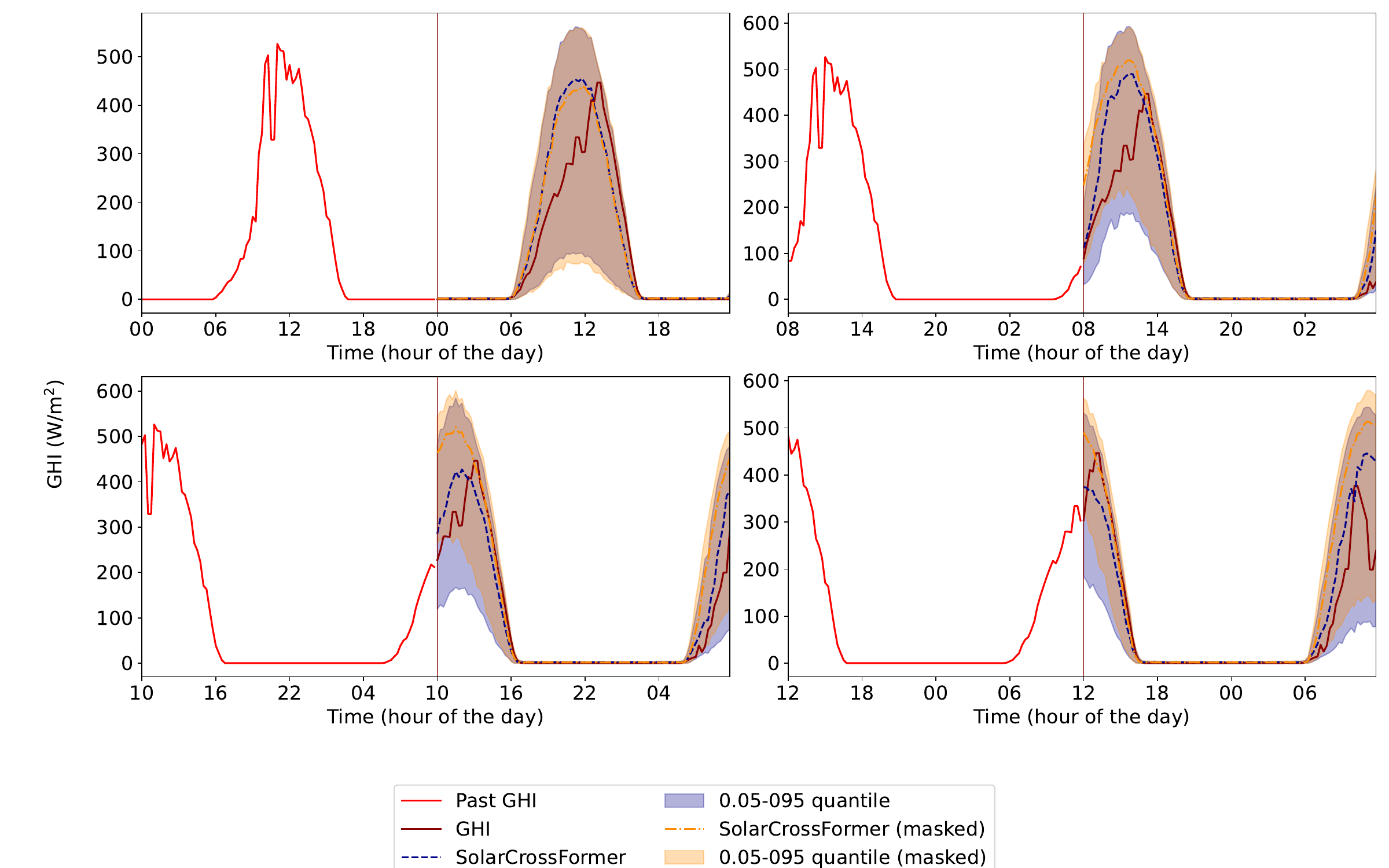} 
    \caption{Comparison of forecasted trajectory in October at Neuchâtel, with past station data masked or unmasked.}
     \label{masked_vs_unmasked_0}
\end{figure}
   \end{flushleft}

\begin{flushleft}
\begin{figure}[h!]
    \includegraphics[width=8.7cm, height=5cm]{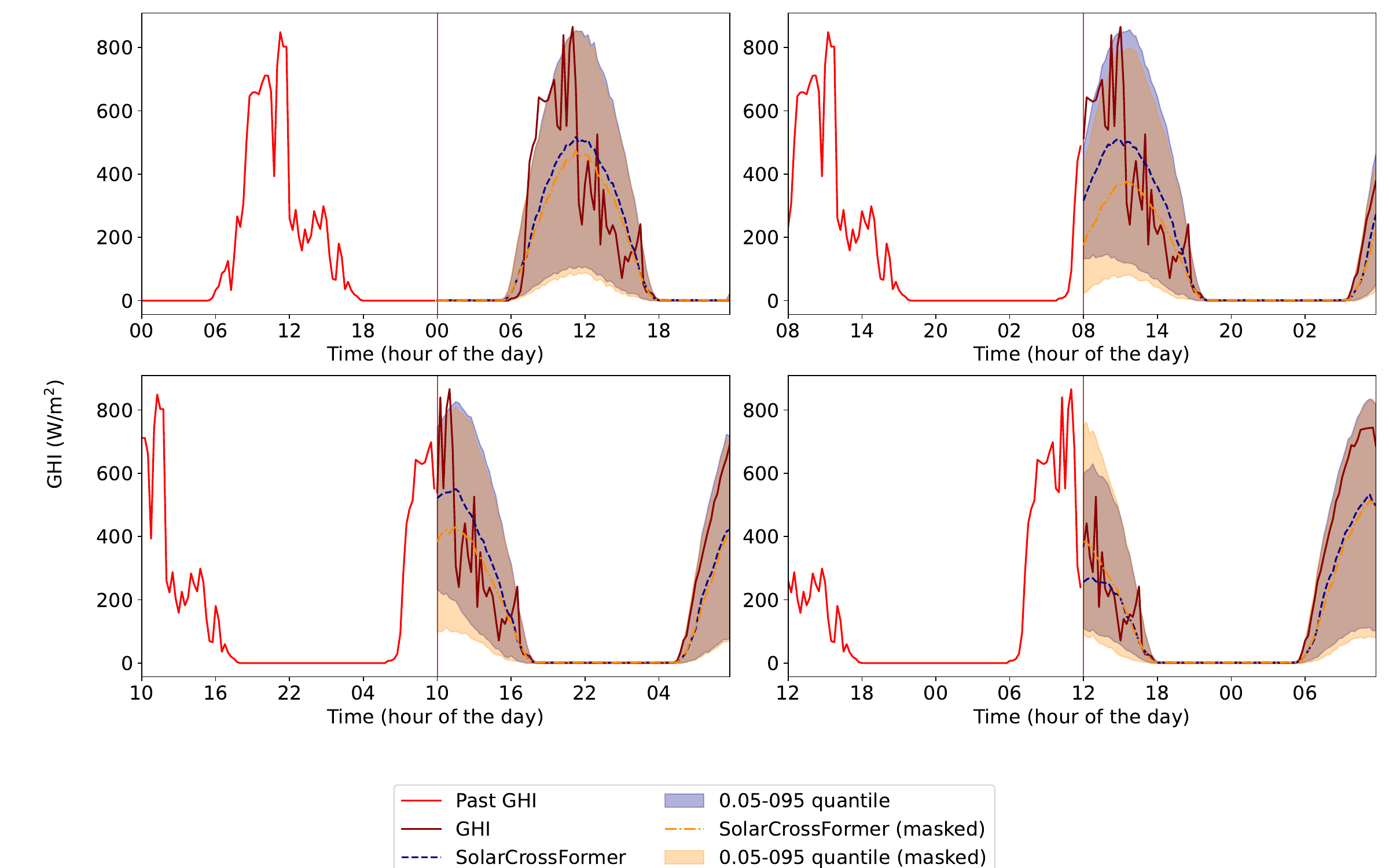} 
    \caption{Comparison of forecasted trajectory in March at Neuchâtel, with past station data masked or unmasked.}
     \label{masked_vs_unmasked_1}
\end{figure}
   \end{flushleft}

In most cases, the degradation of performance between masked and unmasked nodes is relatively low.  On  Figure \ref{comp_with_without_mask} we compare the NMAE and NRMSE over the horizon for the two Swiss nodes, when past station data are entirely masked or entirely available. As expected, the main difference occurs in the short time forecasts, especially the first hour. Afterwards the models accuracy rapidly catches up for Interlaken (where close by nodes with similar conditions are available). Although, a small performance gap is still present for Neuchâtel due to several factors. Among them, we can highlight three factors: the node is located close to the boundary of the node map, Neuchâtel has a micro-climate with high fog frequency in the cold seasons, and neighbouring meteorological stations used in the dataset are placed at higher altitudes where fog is often absent. The short term degradation due to the absence of auto-regression at the predicted node when no past data are available is confirmed by sample trajectories visualization, see Figures \ref{masked_vs_unmasked_0} and \ref{masked_vs_unmasked_1} for days in October and March, respectively.

\section{Conclusion}
\label{section_5}
We have advanced day-ahead irradiance forecasting with SolarCrossformer, a novel deep learning model that fuses information from satellite images with measurements from a network of weather sensors. The main advantages of SolarCrossformer are its robustness and flexibility, which enable deployment in real-life operations when it is needed to incorporate data from unseen locations without retraining the entire model. Furthermore, the model can generate forecasts for locations without any input measurements, utilizing their geographic coordinates, satellite data and the existing sensor network. We have performed an extensive evaluation against state-of-the-art approaches and a commercial NWP solution over a dataset of one full year and 127 locations distributed across Switzerland. Our proposed model, SolarCrossfoemer, has demonstrated improvement in the forecasting accuracy and robustness when forecasting irradiance on locations without historical data. An interesting research direction is to explore the gains of fusing information from other data sources that capture the local weather patterns, such as public cameras \cite{niu2025solar, SARKIS2024112600} or PV power production \cite{Carrillo2023}, with satellite imaging data and coarse NWP that capture the wider spatial context. The mix of weather sensors, public cameras and PV power production can create a denser spatial network of sensors that captures the local irradiance patterns.

\section*{Acknowledgments}
We would like to thank Adib Mellah for preparing and curating part of the datasets used in this study. This research was co-funded by the European Union from the European Union’s Horizon Europe Research and Innovation Programme under Grant Agreement No 101146883 - Project SUPERNOVA.
Views and opinions expressed are however those of the author(s) only and do not necessarily reflect those of the European Union or CINEA. Neither the European Union nor the granting authority can be held responsible for them.


\section{Appendix: Forecasts visualization}

\subsection{Comparison of different deep learning models}

   \begin{flushleft}
\begin{figure}[h!]
    \includegraphics[width=8.7cm, height=5cm]{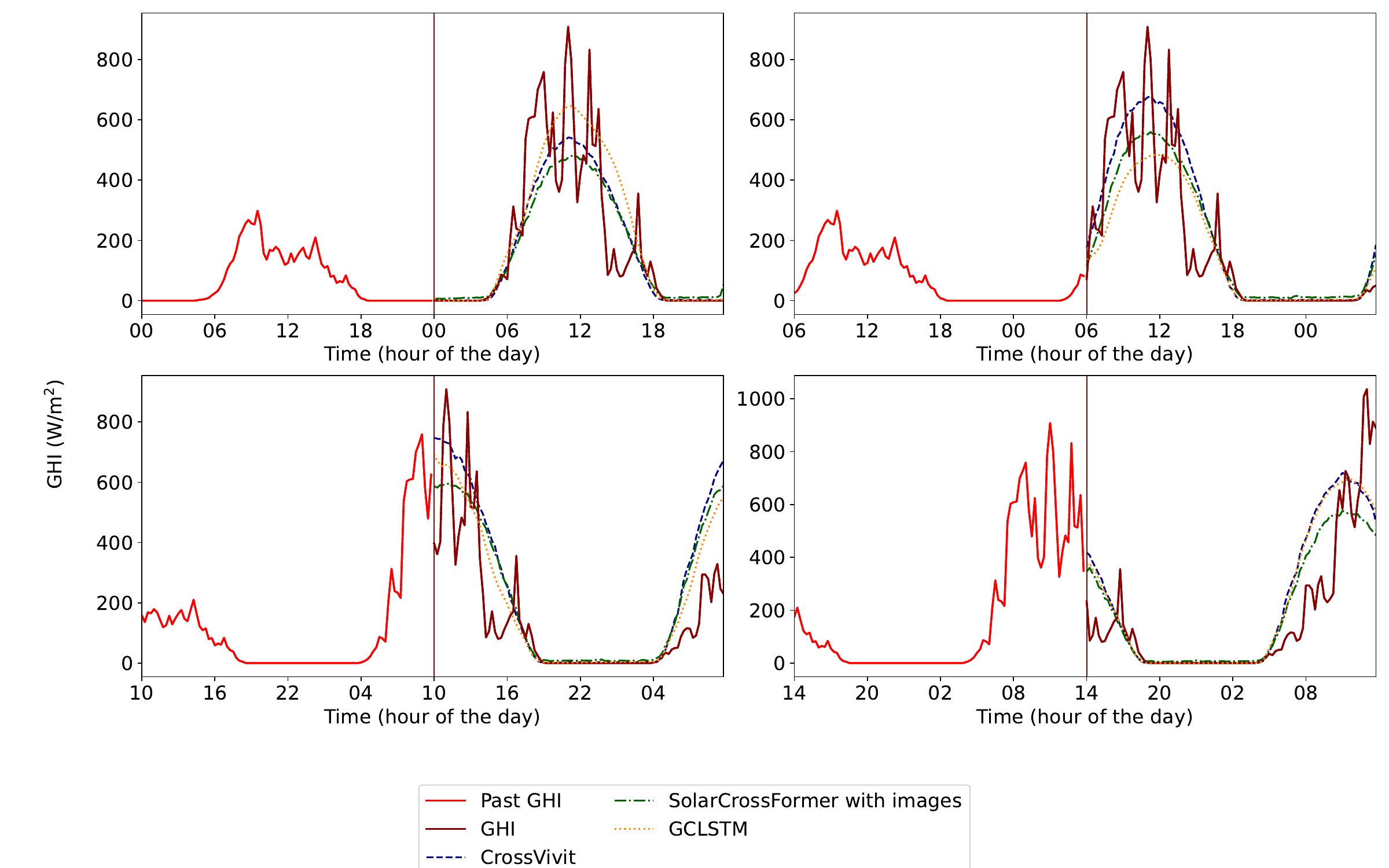} 
    \caption{Comparison of models forecasts (trained under MSE loss) in Bern, for a day with varying cloudiness level in May  2024.}
     \label{mod_comp_1}
\end{figure}
   \end{flushleft}

   \begin{flushleft}
\begin{figure}[h!]
    \includegraphics[width=8.7cm, height=5cm]{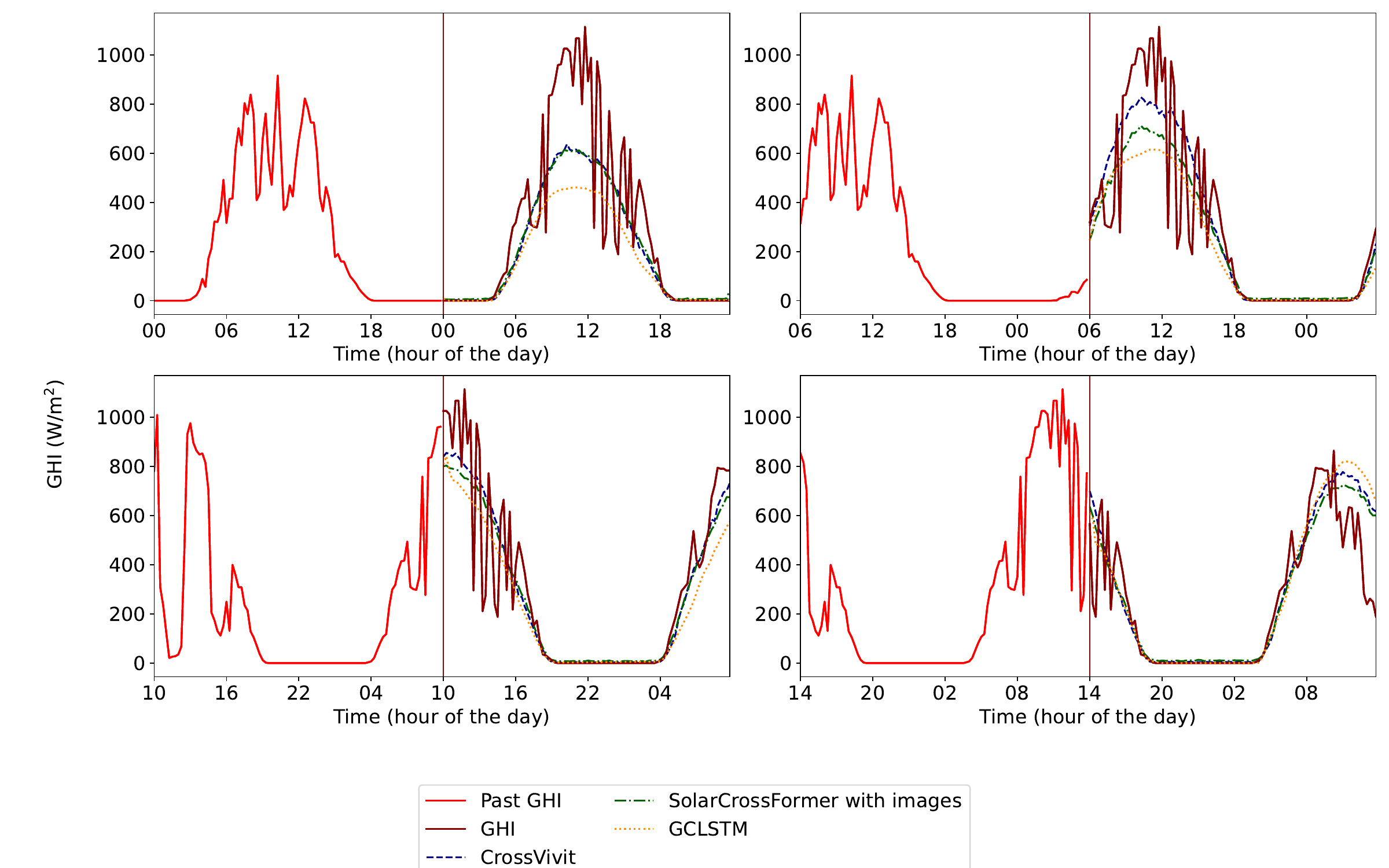} 
    \caption{Comparison of models forecasts (trained under MSE loss) in Bern, for a day with varying cloudiness level in May 2024.}
     \label{mod_comp_2}
\end{figure}
   \end{flushleft}

   \begin{flushleft}
\begin{figure}[h!]
    \includegraphics[width=8.7cm, height=5cm]{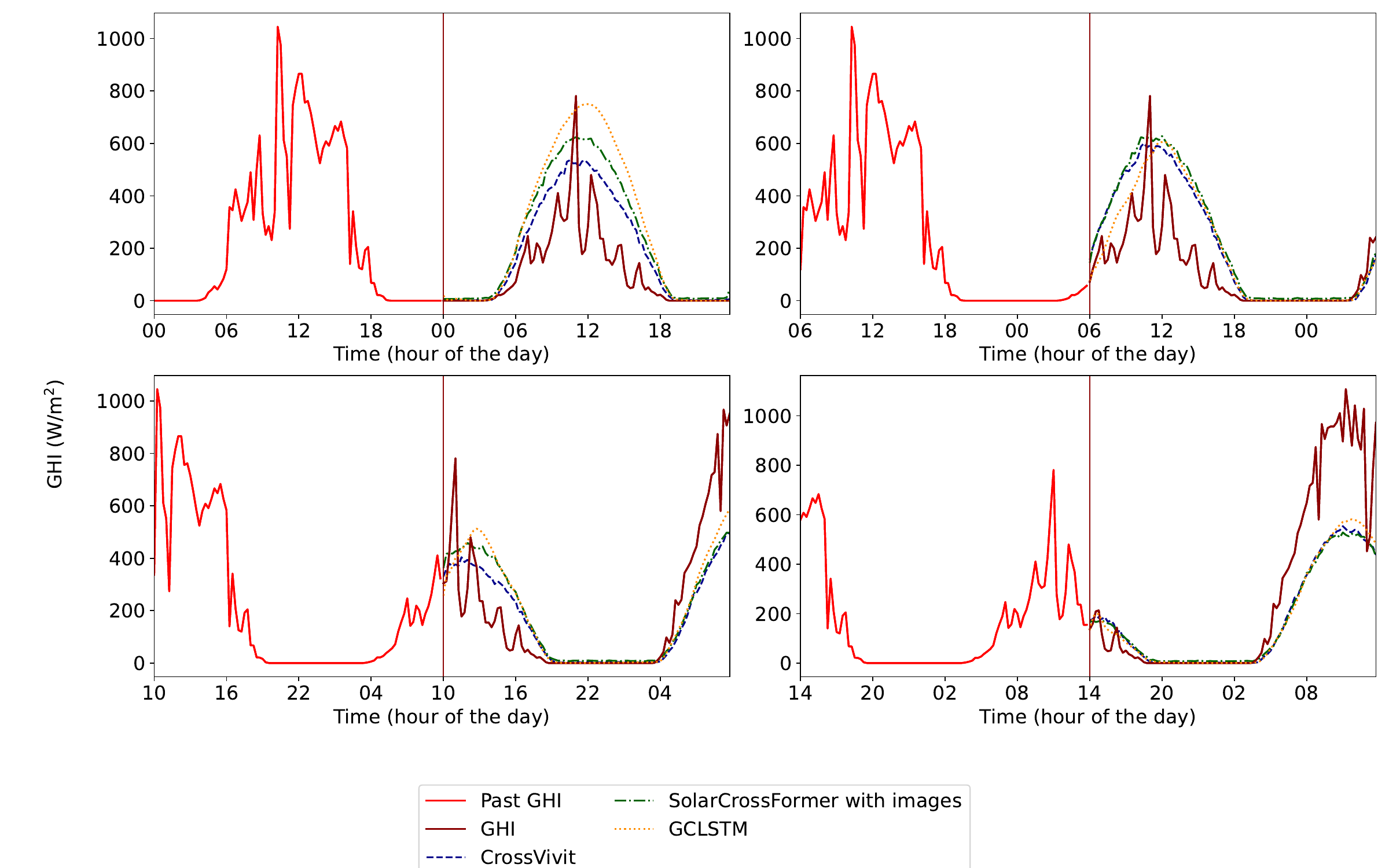} 
    \caption{Comparison of models forecasts (trained under MSE loss)  in Bern, for a day with varying cloudiness level in June 2024.}
     \label{mod_comp_3}
\end{figure}
   \end{flushleft}

\vspace{30pt}

\subsection{NWP-based forecasts vs SolarCrossFormer}
\vspace{30pt}

   \begin{flushleft}
\begin{figure}[h!]
    \includegraphics[width=8.7cm, height=5cm]{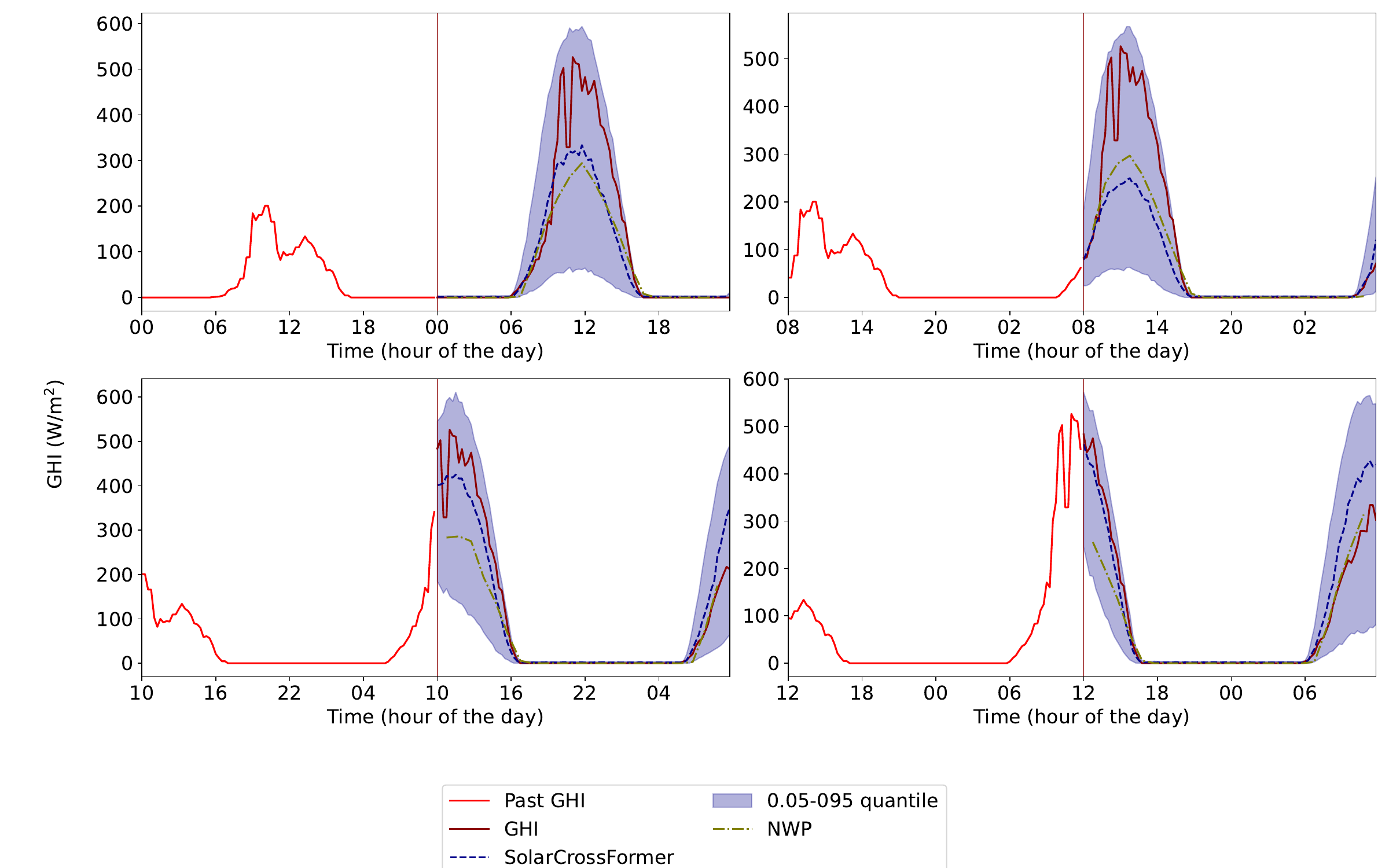} 
    \caption{Comparison of NWP forecasts and SolarCrossFormer forecasts in Neuchâtel, for a sunny day after a foggy day, in October 2024.}
     \label{nwp_vs_solarformer_sunny}
\end{figure}
   \end{flushleft}

\vspace{10pt}
\subsection{Forecasts without station data}

\bibliography{PV_bibliography}
\bibliographystyle{IEEEtran}


%
%
%
%
%

\end{document}